\pdfoutput=1

\documentclass[11pt]{article}

\usepackage[final]{acl}

\usepackage{times}
\usepackage{latexsym}

\usepackage[T1]{fontenc}

\usepackage[utf8]{inputenc}

\usepackage{microtype}

\usepackage{inconsolata}

\usepackage{graphicx}
\usepackage{subcaption}

\usepackage[normalem]{ulem}
\usepackage[dvipsnames]{xcolor}

\newcommand{\tightcolorbox}[2]{%
  {\setlength{\fboxsep}{2pt}%
   \colorbox{#1}{#2}}%
}

\usepackage{booktabs}
\usepackage{multirow}
\usepackage{tabularx}
\usepackage{float}
\usepackage[table]{xcolor}

\usepackage{pifont}
\newcommand{\cmark}{\textcolor{green!60!black}{\ding{51}}}
\newcommand{\xmark}{\textcolor{red}{\ding{55}}}  

\usepackage{colortbl}

\definecolor{g1}{RGB}{218,237,223}  
\definecolor{g2}{RGB}{186,224,196}
\definecolor{g3}{RGB}{148,208,165}
\definecolor{g4}{RGB}{105,191,127}
\definecolor{g5}{RGB}{67,172,93}
\definecolor{g6}{RGB}{42,145,75}
\definecolor{g7}{RGB}{0,109,44}     

\newcommand{\heatcell}[2]{\cellcolor{#1}#2}

\title{Fine-Tuning on Noisy Instructions: Effects on Generalization and Performance}

\author{Ahmed Alajrami \quad Xingwei Tan \quad  Nikolaos Aletras\\
  Department of Computer Science \\
  University of Sheffield, UK \\
  \texttt{ \small \{ajsalajrami1, xingwei.tan, n.aletras\}@sheffield.ac.uk}
  }

\begin{document}
\maketitle
\begin{abstract}
Instruction-tuning plays a vital role in enhancing the task-solving abilities of large language models (LLMs), improving their usability in generating helpful responses on various tasks. However, previous work has demonstrated that they are sensitive to minor variations in instruction phrasing. In this paper, we explore whether introducing perturbations in instruction-tuning data can enhance LLMs' resistance against noisy instructions. We focus on how instruction-tuning with perturbations, such as removing stop words or shuffling words, affects LLMs' performance on the original and perturbed versions of widely-used benchmarks (MMLU, BBH, GSM8K). We further assess learning dynamics and potential shifts in model behavior. Surprisingly, our results suggest that instruction-tuning on perturbed instructions can, in some cases, improve downstream performance.
These findings highlight the importance of including perturbed instructions in instruction-tuning, which can make LLMs more resilient to noisy user inputs.\footnote{Code is available here: \url{https://github.com/aajrami/finetuning-on-noisy-instructions/}}
\end{abstract}

\section{Introduction}

Instruction-tuning is widely adopted to enable LLMs to follow complex instructions and respond properly~\citep{sanh2022multitask,zhao2023survey,Chang2023ASO,minaee2024large,zhang2024instructiontuninglargelanguage}.
During instruction-tuning, LLMs are fine-tuned on datasets comprising various task instructions and their corresponding responses. 

\begin{figure}[!t]
    \centering
    \includegraphics[width=0.99\linewidth]{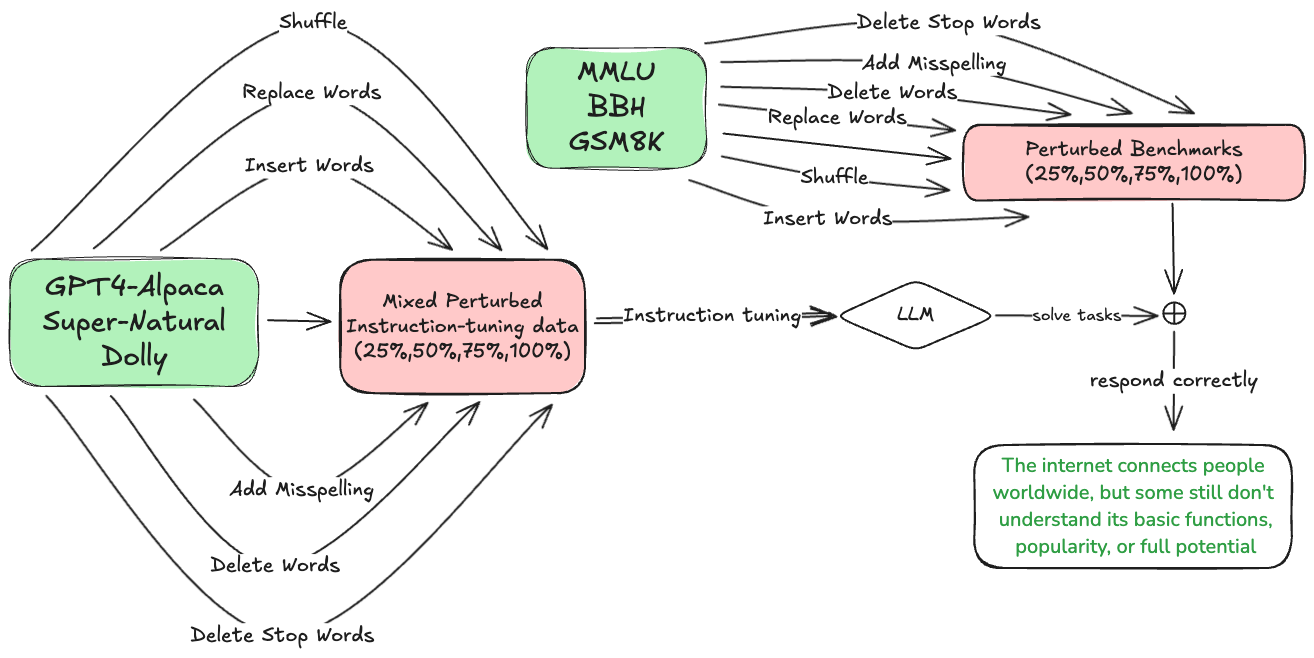}
    \caption{Instruction-tuning on perturbed instructions can enhance LLM's resilience to noisy inputs.}
    \label{fig:motivation}
\end{figure}

LLMs have been shown to be sensitive to prompt variability, producing inconsistent responses when given semantically equivalent prompts \cite{sun2024evaluating,zhao-etal-2024-improving,yan-etal-2024-contrastive}. To remedy this, recent instruction datasets are often generated with extensive paraphrasing using LLMs to increase data diversity \citep{peng2023instruction}. However, this paraphrased data is of high quality with minimal noise in the instructions.
A different line of work has explored the robustness of instruction-tuned models to instruction variations during \textit{inference} by introducing different types of noise, such as deleting words \citep{gu2023robustness}. However, how noisy data may affect LLMs during \textit{training} has yet to be explored.

In this paper, we focus on answering the following research question: 
\textit{Can fine-tuning of base models on perturbed instructions improve their resilience to noisy user inputs?} Our question is theoretically motivated by previous work that has shown that introducing noise during training acts as a form of regularization~\citep{bishop1995training}, which can prevent overfitting and improve generalization.\footnote{We may interchangeably use terms such as `resilience' or `robustness' to refer to the model's capacity to withstand and adapt to noisy user inputs without a substantial drop in performance. The term `generalization' refers to the model's capacity to perform a given task when presented with novel, previously unseen phrasings or formats.}

To evaluate the impact of instruction perturbation and simulate noisy user inputs, we employ five strategies inspired by \citet{gu2023robustness}: (1) delete stop words, (2) shuffle words, (3) delete words, (4) replace words, and (5) insert words. We further introduce a sixth perturbation strategy by adding misspellings. 
These strategies allow us to simulate and analyze various forms of perturbed instructions, including both structural and semantic changes. 
We construct four perturbed instruction-tuning datasets where GPT4-Alpaca~\citep{peng2023instruction}, Super-Natural~\citep{wang2022super}, and Dolly~\citep{conover2023free} are combined and then perturbed.
We include versions of the data where 25\%, 50\%, 75\%, and 100\% of the instructions are perturbed respectively, which allows us to compare how LLMs are affected by different proportions of noisy instructions presented in instruction-tuning.
We evaluate performance across three widely-used language understanding benchmarks: Massive Multitask Language Understanding~\citep{hendrycks2021measuring}, Big-Bench Hard~\citep{Suzgun2022ChallengingBT}, and Grade School Math~\citep{Cobbe2021TrainingVT}.  
Figure~\ref{fig:motivation} shows the process of generating noisy instruction-tuning data and LLM fine-tuning.

\paragraph{Contributions.} We make two key contributions. First, we conduct a systematic study of how noisy instructions, by fundamentally altering their syntactic and semantic structure during training, impact LLM performance on downstream tasks. Second, our empirical analysis suggests that fine-tuning on noisy instructions may offer a simple approach to enhance robustness. The results of our study appear to offer insights into the nature of LLM learning during instruction-tuning. They prompt a re-examination of the widely held assumption that complete instruction comprehension is always necessary for effective task learning. Specifically, our findings suggest that LLMs can derive benefit from instruction modifications that do not strictly preserve meaning, indicating a more nuanced relationship between instruction and task performance than previously assumed.

\begin{table*}[!t]
  \centering
  \scriptsize
  \resizebox{0.9\textwidth}{!}{%
  \small{
    \begin{tabularx}{1.02\textwidth}{lX}
    \toprule
\multicolumn{1}{c}{\textbf{Perturbation}} & \multicolumn{1}{c}{\textbf{User's Content}} \\
\hline
    \multirow{2}{*}{No (Original)} & \texttt{<instruction>} Rewrite the given paragraph in a shorter, easier to understand form. \texttt{<\textbackslash instruction>}
    \newline Input: Although it is generally accepted that the internet has allowed us to connect with people all \ldots \\
    \hline
    
    \multirow{2}{*}{Delete Stop Words} & \texttt{<instruction>} Rewrite \tightcolorbox{GreenYellow}{\sout{the}} given paragraph \tightcolorbox{GreenYellow}{\sout{in a}} shorter, easier \tightcolorbox{GreenYellow}{\sout{to}} understand form. \texttt{<\textbackslash instruction>}
    \newline Input: Although it is generally accepted that the internet has allowed us to connect with people all \ldots \\
    \hline
    
    \multirow{2}{*}{Shuffle Words} & \texttt{<instruction>} Rewrite \tightcolorbox{GreenYellow}{shorter} given paragraph in a \tightcolorbox{GreenYellow}{easier}, \tightcolorbox{GreenYellow}{the} to understand form.\texttt{<\textbackslash instruction>}
    \newline Input: Although it is generally accepted that the internet has allowed us to connect with people all \ldots \\
    \hline
    
    \multirow{2}{*}{Delete Words} & \texttt{<instruction>} Rewrite the given \tightcolorbox{GreenYellow}{\sout{paragraph}} \tightcolorbox{GreenYellow}{\sout{in}} a shorter, easier \tightcolorbox{GreenYellow}{\sout{to}} understand form. \texttt{<\textbackslash instruction>}
    \newline Input: Although it is generally accepted that the internet has allowed us to connect with people all \ldots \\
    \hline
    
    \multirow{2}{*}{Replace Words} & \texttt{<instruction>} Rewrite the \tightcolorbox{GreenYellow}{previous} paragraph in a \tightcolorbox{GreenYellow}{new}, easier to understand \tightcolorbox{GreenYellow}{it}. \texttt{<\textbackslash instruction>}
    \newline Input: Although it is generally accepted that the internet has allowed us to connect with people all \ldots \\
    \hline
    
    \multirow{3}{*}{Insert Words} & \texttt{<instruction>} Rewrite the given paragraph in a shorter \tightcolorbox{GreenYellow}{form}, easier \tightcolorbox{GreenYellow}{than} to understand form \tightcolorbox{GreenYellow}{better}. \texttt{<\textbackslash instruction>}
    \newline Input: Although it is generally accepted that the internet has allowed us to connect with people all ... \\
    \hline
    
    \multirow{2}{*}{Add Misspelling} & \texttt{<instruction>} Rewrite the \tightcolorbox{GreenYellow}{givdn paragraphu} in a shorter, easier to understand \tightcolorbox{GreenYellow}{frm}. \texttt{<\textbackslash instruction>}
    \newline Input: Although it is generally accepted that the internet has allowed us to connect with people all \ldots \\
    
    \bottomrule
    \end{tabularx}%
}}
\caption{Example of a task instruction from the GPT4-Alpaca dataset and the corresponding perturbed instruction generated by each perturbation strategy.}
\label{tab:perturbation-examples}
\end{table*}


\section{Related Work}
\subsection{Analyzing Instruction-tuning}
Instruction fine-tuning enables LLMs to follow user instructions and reduces the need for few-shot in-context examples~\cite{ouyang2022training,wei2022finetuned,touvron2023llamaopenefficientfoundation,chung2024scaling}. 
The instruction-tuning datasets contain instructions of various tasks and their corresponding responses which can be human-annotated \cite{mishra-etal-2022-cross} or synthetically generated \cite{alpaca,peng2023instruction}.

AutoPrompt \cite{shin-etal-2020-autoprompt}, which applied a gradient-based search to optimize the prompt for various tasks, usually finds prompts that are hardly comprehensible by humans, indicating that language models have a vastly different way to understand instructions.
Recent studies have investigated the internal mechanisms of instruction fine-tuning and their influence on LLMs. 
By observing the output token distribution shift of models before and after instruction-tuning, \citet{lin2024the} found that most shifts occur with stylistic tokens (e.g. discourse markers and transitional words), and knowledge content originates from untuned LLMs.
By introducing knowledge interventions, \citet{ren-etal-2024-learning} also showed that instruction-tuning is a process of self-aligning the instructions with existing parametric knowledge rather than introducing new knowledge into the model.

\subsection{Robustness of Instruction-tuned Models}
\citet{gu2023robustness} investigated how instruction-tuned models handle instruction perturbations and paraphrasing. After fine-tuning a model on the original instructions training set and evaluating it on the perturbed instructions test set, they found the model was relatively robust in few-shot settings but notably sensitive in zero-shot scenarios. Similarly, \citet{sun2024evaluating} showed that paraphrased instructions can disrupt model consistency and proposed a mitigation strategy using soft prompt embeddings to align semantically similar instructions. \citet{wang-etal-2024-resilience} examined errors from speech recognition and OCR, finding such noise significantly degrades LLMs performance.
They also explored using LLMs for zero-shot correction of noisy instructions.
In addition, \citet{abedin2025arithmattack} demonstrated that LLMs are sensitive to noise in reasoning tasks by introducing random punctuation perturbations into math problem prompts.

\citet{yan-etal-2024-contrastive} proposed contrastive instruction-tuning which align the hidden representations of instruction-instance pairs that are semantically equivalent but textually different while distinguishing those that are semantically different.
\citet{zhao-etal-2024-improving} proposed a consistency alignment framework that incorporates instruction augmentation through paraphrasing and automatic self-reward mechanisms.
\citet{lou2024muffin} introduced a dataset curation scheme that diversifies task inputs across multiple facets to enhance instruction diversity.
\citet{kim2024instructive} presented instructive decoding, a method that strengthens instruction-following abilities in instruction-tuned LLMs by contrasting decoding paths without additional fine-tuning.

\citet{luo2024robustft} introduced a denoising framework that detects and filters out low-quality samples, enabling more robust fine-tuning of models on noisy downstream datasets. Similarly, \citet{agrawal2025enhancing} performed an empirical analysis focused on enhancing model robustness to character- and word-level perturbations in classification tasks, revealing that iterative self-denoising surpasses approaches like ensembling and representation alignment.

Unlike previous studies, we investigate how \textit{training} on noisy instructions affects their ability to adapt to instruction perturbations.

\section{Instruction Perturbation Strategies}

\label{perturbation_methodolgies}

We investigate the impact of instruction fine-tuning on the performance of LLMs when subjected to noisy input conditions.
Following \citet{gu2023robustness}, we employ five instruction perturbation strategies: delete stop words, shuffle words, delete words, replace words, and insert words. 
Furthermore, we introduce misspelling as an additional noise injection approach. 
Table \ref{tab:perturbation-examples} shows an example instruction from the GPT4-Alpaca dataset~\citep{peng2023instruction}. 
The model input consists of an instruction and associated context. The perturbation strategies are applied exclusively to the instruction component.

\paragraph{Delete Stop Words.}

Stop words, such as ``the'', ``is'', and ``of'' are functional words that mainly contribute to grammatical structure but have limited effects on semantic content. 
Removing stop words leads to syntactically incomplete instructions, allowing us to evaluate the model's reliance on syntactic cues and its ability to infer meaning from partial input.
For instance, the instruction \textit{Translate the sentence into French} becomes \textit{Translate sentence French}.

\paragraph{Shuffle Words.}

The second perturbation strategy is to randomly shuffle the words.
By changing the original word order, we aim to introduce both syntactic and semantic alterations. 
We employ a 25\% word shuffling of words within the instruction, while the other words maintain their relative order. We cap shuffling at 25\%, enough to mimic the partial mix-up would be seen from hurried typing, and realistic enough to test robustness without turning the prompt into total gibberish.
For instance, given the instruction \textit{Summarize the following paragraph}, this might result in \textit{following the Summarize paragraph}.
This perturbation provides an assessment of how sensitive models are to changes in word order and to what extent they rely on the original structure to comprehend the instruction.

\paragraph{Delete Words.}
In this perturbation strategy, 25\% of the words within each instruction are randomly deleted. In contrast to the targeted removal of stop words, this approach introduces more significant distortions by potentially eliminating both functional and semantic words, thereby disrupting the semantic coherence of the instructions to a greater extent. This strategy assesses the model's capacity to infer task intent when presented with incomplete syntactic and semantic structures.

\paragraph{Replace Words.}
We randomly select 25\% of the words from an instruction and replace them using predictions from a pretrained BERT model~\citep{devlin-etal-2019-bert}.
Following \citet{gu2023robustness}, we use BERT's masked language modeling head to generate a contextually plausible substitute for all selected words.
The selected words are replaced by \texttt{[MASK]} tokens, then BERT predicts replacement words in a forward pass.
This strategy introduces minimal semantic shifts to the instructions without relying on a lexicon, which typically requires manual effort.
This perturbation may or may not alter the core meaning of the instruction, depending on the replaced words and their context.
The strategy allows us to investigate the model's sensitivity to nuanced lexical variations and its ability to generalize under slightly altered task phrasing.

\paragraph{Insert Words.}
We introduce additional words into the instruction by leveraging a pretrained BERT model~\citep{devlin-etal-2019-bert}.
Specifically, we randomly select positions between existing words, covering approximately 25\% of the total word count, and insert a \texttt{[MASK]} token at each selected position.
We then replace each \texttt{[MASK]} with words predicted by BERT, resulting in an augmented instruction containing additional, contextually plausible tokens.
This perturbation may introduce noise, redundancy, or shifts in meaning, challenging the model's ability to extract the core task intention from a more verbose or distorted input.

\paragraph{Add Misspelling.}
Finally, to simulate noisy input that may more closely resemble user-generated errors, we introduce typographical errors. We randomly select 25\% of the words within each instruction and introduce a typo into each of these words. We apply simple character-level edits, such as deleting a random letter, transposing adjacent letters, inserting a random vowel, or substituting a character with a randomly selected one. This strategy enables us to evaluate the model's sensitivity to spelling errors in the instructions and its ability to discern the intended meaning from noisy input.

\section{Experimental Setup}

\subsection{Models}
We experiment with two open-weight base LLMs in two sizes:
Qwen-2.5 (7B and 72B) \citep{yang2024qwen2}; and Llama-3.1 (8B and 70B) \citep{dubey2024llama}.

\renewcommand*{\arraystretch}{1.1}
\begin{table}[!t]
\small
\begin{center}
\begin{tabular}{lcc}
\toprule
\textbf{Instruction Dataset} & \textbf{\# Samples}  \\ \midrule
GPT4-Alpaca & 52,002 \\ 
Super-Natural Instruction & 55,793 \\ 
Dolly & 15,011 \\ \midrule
Total & 122,806 \\
\bottomrule
\end{tabular}
\caption{Number of samples in each dataset.} 
\label{table:datasets_size}
\end{center}
\end{table}

\subsection{Instruction Datasets}
We fine-tune all the base models using a combination of three standard instruction datasets with distinct characteristics. 

\textbf{GPT4-Alpaca} \citep{peng2023instruction} is derived from Alpaca \citep{alpaca}, where the original examples are replaced with responses generated by GPT-4. \textbf{Super-Natural Instruction} \citep{wang2022super} contains diverse tasks, including text classification and translation, with corresponding instructions. It is designed to evaluate the LLM abilities across a wide range of linguistic and functional contexts. \textbf{Dolly} \citep{conover2023free} consists of instruction-following examples that reflect practical, real-world tasks like brainstorming and creative writing. The prompt-response pairs are high-quality and human-generated. Table \ref{table:datasets_size} summarizes the number of samples in these datasets.

\paragraph{Perturbation Settings.}
To simulate real-world settings where the perturbations could appear altogether, we construct five different dataset mixtures, each containing a different proportion of perturbed instructions:
(1) the original, unmodified instruction samples from all three datasets considered as a baseline (\textbf{0\% Perturbation}), (2) 25\% of the instruction samples are perturbed, while the remaining 75\% are left unaltered (\textbf{25\% Perturbation}), (3) half of the instruction samples are perturbed (\textbf{50\% Perturbation}), (4) 75\% of the samples are perturbed (\textbf{75\% Perturbation}), and (5) all instruction samples across the three datasets are perturbed (\textbf{100\% Perturbation}).

In all mixtures involving perturbations, the altered samples are evenly distributed across the six different perturbation strategies (Section \ref{perturbation_methodolgies}).

\subsection{Implementation Details}
We apply parameter-efficient fine-tuning methods for all experiments. Specifically, we use LoRA \citep{hu2022lora} to fine-tune the 7B and 8B models, and QLoRA \citep{dettmers2023qlora} for the larger 70B and 72B models.
Each model is fine-tuned for one epoch on each dataset mixture to ensure consistency across experiments. All fine-tuning runs were performed on a single NVIDIA H100 GPU.
Full details on the fine-tuning hyperparameters are provided in Appendix \ref{sec:appendix_hyperparameters}.


\begin{table*}[!t]
\begin{center}
\small
\resizebox{\textwidth}{!}{%
\begin{tabular}{clccccc|ccccc|ccccc}
\toprule
& & \multicolumn{5}{c}{\textbf{MMLU (5-shot)}} & \multicolumn{5}{c}{\textbf{BBH (CoT)}} & \multicolumn{5}{c}{\textbf{GSM8K (CoT)}} \\

\textbf{} & \textbf{IT} & \textsc{0\%} & \textsc{25\%} & \textsc{50\%} & \textsc{75\%} & \textsc{100\%} & \textsc{0\%} & \textsc{25\%} & \textsc{50\%} & \textsc{75\%} & \textsc{100\%} & \textsc{0\%} & \textsc{25\%} & \textsc{50\%} & \textsc{75\%} & \textsc{100\%} \\

\midrule

\multirow{6}{*}{\rotatebox{90}{\footnotesize Qwen 7B}}
& \textsc{van} & 74.3 \textsubscript{0.0} & 73.0 \textsubscript{0.1} & 71.5 \textsubscript{0.1} & 70.0 \textsubscript{0.4} & 68.6 \textsubscript{0.6}
 & 66.7 \textsubscript{0.1} & 63.9 \textsubscript{0.4} & 60.8 \textsubscript{0.4} & 57.7 \textsubscript{0.5} & 54.9 \textsubscript{0.5}
 & 79.9 \textsubscript{0.2} & 12.5 \textsubscript{0.3} & 22.9 \textsubscript{0.6} & 33.0 \textsubscript{1.4} & 42.7 \textsubscript{1.2}

 \\ 
 \cmidrule{2-17}
& \textsc{0\%} & \heatcell{g1}74.3 \textsubscript{0.0} & \heatcell{g1}73.0 \textsubscript{0.1} & \heatcell{g1}71.7 \textsubscript{0.1} & \heatcell{g1}70.2 \textsubscript{0.4} & \heatcell{g1}68.9 \textsubscript{0.7}
 & \heatcell{g3}66.8 \textsubscript{0.0} & \heatcell{g1}62.7 \textsubscript{0.2} & \heatcell{g1}58.7 \textsubscript{0.2} & \heatcell{g1}54.3 \textsubscript{0.5} & \heatcell{g1}50.6 \textsubscript{0.6}
 & \heatcell{g4}80.6 \textsubscript{0.0} & \heatcell{g3}12.6 \textsubscript{0.5} & \heatcell{g2}24.5 \textsubscript{0.5} & \heatcell{g5}\textbf{34.6} \textsubscript{1.0} & \heatcell{g4}44.6 \textsubscript{1.2}

 \\ 
& \textsc{25\%} & \heatcell{g5}\textbf{74.4} \textsubscript{0.0} & \heatcell{g1}73.0 \textsubscript{0.1} & \heatcell{g2}71.8 \textsubscript{0.1} & \heatcell{g2}70.3 \textsubscript{0.5} & \heatcell{g2}69.1 \textsubscript{0.6}
 & \heatcell{g2}66.7 \textsubscript{0.0} & \heatcell{g2}63.3 \textsubscript{0.3} & \heatcell{g2}59.7 \textsubscript{0.2} & \heatcell{g2}55.9 \textsubscript{0.5} & \heatcell{g2}52.4 \textsubscript{0.6}
 & \heatcell{g5}\textbf{81.1} \textsubscript{0.0} & \heatcell{g3}12.6 \textsubscript{0.4} & \heatcell{g4}24.7 \textsubscript{0.5} & \heatcell{g4}34.5 \textsubscript{1.2} & \heatcell{g1}44.2 \textsubscript{1.4}

 \\ 
& \textsc{50\%} & \heatcell{g5}\textbf{74.4} \textsubscript{0.0} & \heatcell{g5}\textbf{73.1} \textsubscript{0.1} & \heatcell{g5}\textbf{71.9} \textsubscript{0.1} & \heatcell{g5}\textbf{70.5} \textsubscript{0.5} & \heatcell{g2}69.1 \textsubscript{0.7}
 & \heatcell{g4}67.0 \textsubscript{0.0} & \heatcell{g5}\textbf{64.0} \textsubscript{0.2} & \heatcell{g5}\textbf{61.1} \textsubscript{0.3} & \heatcell{g5}\textbf{57.7} \textsubscript{0.6} & \heatcell{g5}\textbf{54.8} \textsubscript{0.6}
 & \heatcell{g4}80.6 \textsubscript{0.0} & \heatcell{g3}12.6 \textsubscript{0.5} & \heatcell{g3}24.6 \textsubscript{0.5} & \heatcell{g3}34.3 \textsubscript{0.7} & \heatcell{g3}44.5 \textsubscript{1.1}

 \\ 
& \textsc{75\%} & \heatcell{g1}74.3 \textsubscript{0.0} & \heatcell{g1}73.0 \textsubscript{0.1} & \heatcell{g2}71.8 \textsubscript{0.2} & \heatcell{g4}70.4 \textsubscript{0.5} & \heatcell{g2}69.1 \textsubscript{0.7}
 & \heatcell{g5}\textbf{67.4} \textsubscript{0.0} & \heatcell{g4}63.9 \textsubscript{0.3} & \heatcell{g4}60.7 \textsubscript{0.2} & \heatcell{g4}57.6 \textsubscript{0.4} & \heatcell{g4}54.7 \textsubscript{0.6}
 & \heatcell{g3}80.5 \textsubscript{0.0} & \heatcell{g5}\textbf{12.8} \textsubscript{0.5} & \heatcell{g1}24.4 \textsubscript{0.6} & \heatcell{g2}34.0 \textsubscript{0.9} & \heatcell{g2}44.4 \textsubscript{0.9}

 \\ 
& \textsc{100\%} & \heatcell{g1}74.3 \textsubscript{0.0} & \heatcell{g5}\textbf{73.1} \textsubscript{0.0} & \heatcell{g5}\textbf{71.9} \textsubscript{0.1} & \heatcell{g5}\textbf{70.5} \textsubscript{0.5} & \heatcell{g5}\textbf{69.2} \textsubscript{0.6}
 & \heatcell{g1}66.6 \textsubscript{0.0} & \heatcell{g3}63.4 \textsubscript{0.2} & \heatcell{g3}60.3 \textsubscript{0.3} & \heatcell{g3}56.8 \textsubscript{0.4} & \heatcell{g3}53.8 \textsubscript{0.7}
 & \heatcell{g2}80.0 \textsubscript{0.0} & \heatcell{g3}12.6 \textsubscript{0.2} & \heatcell{g5}\textbf{24.8} \textsubscript{0.6} & \heatcell{g3}34.3 \textsubscript{1.1} & \heatcell{g5}\textbf{45.1} \textsubscript{0.8}

 \\ 

\midrule

\multirow{6}{*}{\rotatebox{90}{\footnotesize Llama 8B}}
& \textsc{van} & 65.8 \textsubscript{0.0} & 64.5 \textsubscript{0.1} & 63.1 \textsubscript{0.1} & 62.1 \textsubscript{0.3} & 60.8 \textsubscript{0.5}
 & 64.5 \textsubscript{0.1} & 62.5 \textsubscript{0.3} & 60.2 \textsubscript{0.4} & 57.5 \textsubscript{1.0} & 55.0 \textsubscript{0.9}
 & 56.3 \textsubscript{0.3} & 9.0 \textsubscript{0.7} & 16.3 \textsubscript{0.6} & 23.5 \textsubscript{0.6} & 30.5 \textsubscript{1.2}

 \\ 
 \cmidrule{2-17}
& \textsc{0\%} & \heatcell{g2}65.8 \textsubscript{0.0} & \heatcell{g1}64.6 \textsubscript{0.2} & \heatcell{g1}63.3 \textsubscript{0.1} & \heatcell{g1}62.2 \textsubscript{0.2} & \heatcell{g1}60.7 \textsubscript{0.5}
 & \heatcell{g3}63.0 \textsubscript{0.4} & \heatcell{g2}63.4 \textsubscript{0.1} & \heatcell{g2}61.1 \textsubscript{0.3} & \heatcell{g1}58.7 \textsubscript{0.7} & \heatcell{g3}56.5 \textsubscript{0.6}
 & \heatcell{g4}58.4 \textsubscript{0.0} & \heatcell{g3}9.2 \textsubscript{0.1} & \heatcell{g1}16.6 \textsubscript{0.7} & \heatcell{g2}23.8 \textsubscript{1.0} & \heatcell{g5}\textbf{28.1} \textsubscript{1.0}

 \\ 
& \textsc{25\%} & \heatcell{g4}65.9 \textsubscript{0.0} & \heatcell{g5}\textbf{64.8} \textsubscript{0.3}  & \heatcell{g2}63.4 \textsubscript{0.2} & \heatcell{g2}62.3 \textsubscript{0.2} & \heatcell{g2}60.9 \textsubscript{0.7}
 & \heatcell{g4}66.0 \textsubscript{0.1} & \heatcell{g1}60.5 \textsubscript{0.4} & \heatcell{g1}60.0 \textsubscript{1.9} & \heatcell{g3}59.1 \textsubscript{0.4} & \heatcell{g3}56.5 \textsubscript{0.6}
 & \heatcell{g5}\textbf{58.5} \textsubscript{0.0} & \heatcell{g5}\textbf{9.4} \textsubscript{0.2} & \heatcell{g2}16.8 \textsubscript{0.8} & \heatcell{g3}23.9 \textsubscript{0.7} & \heatcell{g2}27.7 \textsubscript{0.9}

 \\ 
& \textsc{50\%} & \heatcell{g4}65.9 \textsubscript{0.0} & \heatcell{g5}\textbf{64.8} \textsubscript{0.2}  & \heatcell{g4}63.6 \textsubscript{0.1} & \heatcell{g5}\textbf{62.5} \textsubscript{0.2}  & \heatcell{g4}61.0 \textsubscript{0.6}
 & \heatcell{g1}62.7 \textsubscript{0.0} & \heatcell{g5}\textbf{64.4} \textsubscript{0.3} & \heatcell{g5}\textbf{62.0} \textsubscript{0.5} & \heatcell{g5}\textbf{59.3} \textsubscript{0.5} & \heatcell{g4}56.7 \textsubscript{0.5}
 & \heatcell{g3}58.2 \textsubscript{0.0} & \heatcell{g3}9.2 \textsubscript{0.2} & \heatcell{g3}16.9 \textsubscript{0.6} & \heatcell{g5}\textbf{24.0} \textsubscript{0.9} & \heatcell{g3}27.8 \textsubscript{1.0}

 \\ 
& \textsc{75\%} & \heatcell{g1}65.7 \textsubscript{0.0} & \heatcell{g4}64.7 \textsubscript{0.2} & \heatcell{g4}63.6 \textsubscript{0.1} & \heatcell{g5}\textbf{62.5} \textsubscript{0.2}  & \heatcell{g5}\textbf{61.2} \textsubscript{0.5} 
 & \heatcell{g2}62.9 \textsubscript{0.4} & \heatcell{g3}64.1 \textsubscript{0.3} & \heatcell{g3}61.8 \textsubscript{0.3} & \heatcell{g3}59.1 \textsubscript{0.5} & \heatcell{g2}56.3 \textsubscript{0.5}
 & \heatcell{g1}57.4 \textsubscript{0.0} & \heatcell{g1}9.1 \textsubscript{0.2} & \heatcell{g3}16.9 \textsubscript{0.7} & \heatcell{g2}23.8 \textsubscript{0.8} & \heatcell{g1}27.6 \textsubscript{1.2}

 \\ 
& \textsc{100\% } & \heatcell{g5}\textbf{66.0} \textsubscript{0.0}  & \heatcell{g5}\textbf{64.8} \textsubscript{0.3}  & \heatcell{g5}\textbf{63.7} \textsubscript{0.1}  & \heatcell{g5}\textbf{62.5} \textsubscript{0.3}  & \heatcell{g5}\textbf{61.2} \textsubscript{0.5} 
 & \heatcell{g5}\textbf{66.2} \textsubscript{0.1} & \heatcell{g4}64.2 \textsubscript{0.5} & \heatcell{g4}62.0 \textsubscript{0.5} & \heatcell{g4}59.2 \textsubscript{0.8} & \heatcell{g5}\textbf{56.8} \textsubscript{0.4}
 & \heatcell{g4}58.4 \textsubscript{0.0} & \heatcell{g3}9.2 \textsubscript{0.2} & \heatcell{g5}\textbf{17.1} \textsubscript{0.6} & \heatcell{g1}23.7 \textsubscript{1.2} & \heatcell{g3}27.8 \textsubscript{1.5}

 \\ 

\midrule

\multirow{6}{*}{\rotatebox{90}{\footnotesize Qwen 72B}}
& \textsc{van} & 85.7 \textsubscript{0.0} & 84.5 \textsubscript{0.2} & 83.0 \textsubscript{0.3} & 81.8 \textsubscript{0.3} & 80.3 \textsubscript{0.4}
 & 82.7 \textsubscript{0.1} & 79.2 \textsubscript{0.2} & 75.4 \textsubscript{0.2} & 71.7 \textsubscript{0.4} & 68.1 \textsubscript{0.8}
 & 88.8 \textsubscript{0.2} & 14.9 \textsubscript{0.5} & 28.1 \textsubscript{0.7} & 40.8 \textsubscript{1.0} & 53.0 \textsubscript{1.5}

 \\ 
 \cmidrule{2-17}
& \textsc{0\%} & \heatcell{g5}\textbf{85.8} \textsubscript{0.0}  & \heatcell{g1}84.6 \textsubscript{0.2} & \heatcell{g1}83.1 \textsubscript{0.3} & \heatcell{g1}82.0 \textsubscript{0.2} & \heatcell{g1}80.5 \textsubscript{0.5}
 & \heatcell{g5}\textbf{83.8} \textsubscript{0.1} & \heatcell{g5}\textbf{80.5} \textsubscript{0.2} & \heatcell{g4}77.3 \textsubscript{0.3} & \heatcell{g3}73.8 \textsubscript{0.5} & \heatcell{g5}\textbf{70.8} \textsubscript{1.1}
 & \heatcell{g3}90.0 \textsubscript{0.2} & \heatcell{g2}15.3 \textsubscript{0.4} & \heatcell{g1}29.0 \textsubscript{0.4} & \heatcell{g1}42.7 \textsubscript{1.1} & \heatcell{g1}55.0 \textsubscript{1.7}

\\ 
& \textsc{25\%} & \heatcell{g1}85.7 \textsubscript{0.0} & \heatcell{g1}84.6 \textsubscript{0.3} & \heatcell{g1}83.1 \textsubscript{0.3} & \heatcell{g1}82.0 \textsubscript{0.3} & \heatcell{g1}80.5 \textsubscript{0.6}
 & \heatcell{g5}\textbf{83.8} \textsubscript{0.1} & \heatcell{g4}80.4 \textsubscript{0.2} & \heatcell{g5}\textbf{77.4} \textsubscript{0.4} & \heatcell{g5}\textbf{74.0} \textsubscript{0.8} & \heatcell{g5}\textbf{70.8} \textsubscript{1.0}
 & \heatcell{g1}89.9 \textsubscript{0.2} & \heatcell{g2}15.3 \textsubscript{0.5} & \heatcell{g4}29.6 \textsubscript{0.4} & \heatcell{g4}43.0 \textsubscript{1.1} & \heatcell{g3}55.5 \textsubscript{1.6}

\\ 
& \textsc{50\%} & \heatcell{g1}85.7 \textsubscript{0.0} & \heatcell{g2}84.7 \textsubscript{0.3} & \heatcell{g1}83.1 \textsubscript{0.3} & \heatcell{g1}82.0 \textsubscript{0.3} & \heatcell{g5}\textbf{80.7} \textsubscript{0.6}
 & \heatcell{g2}83.3 \textsubscript{0.1} & \heatcell{g1}80.2 \textsubscript{0.2} & \heatcell{g2}77.2 \textsubscript{0.2} & \heatcell{g1}73.7 \textsubscript{0.6} & \heatcell{g4}70.6 \textsubscript{0.8}
 & \heatcell{g1}89.9 \textsubscript{0.2} & \heatcell{g3}15.4 \textsubscript{0.4} & \heatcell{g2}29.3 \textsubscript{0.5} & \heatcell{g2}42.8 \textsubscript{1.4} & \heatcell{g3}55.5 \textsubscript{1.9}

\\ 
& \textsc{75\%} & \heatcell{g5}\textbf{85.8} \textsubscript{0.0}  & \heatcell{g2}84.7 \textsubscript{0.3} & \heatcell{g5}\textbf{83.2} \textsubscript{0.3}  & \heatcell{g5}\textbf{82.1} \textsubscript{0.3}  & \heatcell{g5}\textbf{80.7} \textsubscript{0.5}
 & \heatcell{g4}83.6 \textsubscript{0.0} & \heatcell{g3}80.3 \textsubscript{0.2} & \heatcell{g2}77.2 \textsubscript{0.3} & \heatcell{g4}73.9 \textsubscript{0.6} & \heatcell{g2}70.4 \textsubscript{0.8}
 & \heatcell{g5}\textbf{90.3} \textsubscript{0.3} & \heatcell{g5}\textbf{15.6} \textsubscript{0.3} & \heatcell{g4}29.6 \textsubscript{0.7} & \heatcell{g3}42.9 \textsubscript{1.7} & \heatcell{g3}55.5 \textsubscript{2.3}

\\ 
& \textsc{100\%} & \heatcell{g5}\textbf{85.8} \textsubscript{0.0}  & \heatcell{g5}\textbf{84.8} \textsubscript{0.2}  & \heatcell{g5}\textbf{83.2} \textsubscript{0.3}  & \heatcell{g5}\textbf{82.1} \textsubscript{0.4} & \heatcell{g3}80.6 \textsubscript{0.6} 
 & \heatcell{g4}83.6 \textsubscript{0.0} & \heatcell{g4}80.4 \textsubscript{0.2} & \heatcell{g4}77.3 \textsubscript{0.2} & \heatcell{g3}73.8 \textsubscript{0.4} & \heatcell{g5}\textbf{70.8} \textsubscript{0.8}
 & \heatcell{g4}90.2 \textsubscript{0.1} & \heatcell{g5}\textbf{15.6} \textsubscript{0.5} & \heatcell{g5}\textbf{29.7} \textsubscript{0.7} & \heatcell{g5}\textbf{43.4} \textsubscript{1.8} & \heatcell{g5}\textbf{55.9} \textsubscript{2.0}

\\ 

\midrule

\multirow{6}{*}{\rotatebox{90}{\footnotesize Llama 70B}}
& \textsc{van} & 75.8 \textsubscript{0.0} & 74.1 \textsubscript{0.1} & 72.2 \textsubscript{0.2} & 70.2 \textsubscript{0.4} & 68.5 \textsubscript{0.4}
 & 78.3 \textsubscript{0.1} & 75.7 \textsubscript{0.2} & 73.3 \textsubscript{0.2} & 70.3 \textsubscript{0.4} & 68.1 \textsubscript{0.4}
 & 80.2 \textsubscript{0.1} & 13.0 \textsubscript{0.3} & 24.1 \textsubscript{0.5} & 34.7 \textsubscript{1.5} & 43.9 \textsubscript{0.9}

\\ 
 \cmidrule{2-17}
& \textsc{0\%} & \heatcell{g4}78.1 \textsubscript{0.0} & \heatcell{g3}76.7 \textsubscript{0.3} & \heatcell{g2}74.9 \textsubscript{0.3} & \heatcell{g2}73.0 \textsubscript{0.4} & \heatcell{g2}71.4 \textsubscript{0.5}
 & \heatcell{g5}\textbf{81.8} \textsubscript{0.1} & \heatcell{g4}78.9 \textsubscript{0.1} & \heatcell{g1}75.9 \textsubscript{0.2} & \heatcell{g1}72.7 \textsubscript{0.4} & \heatcell{g1}70.1 \textsubscript{0.4}
 & \heatcell{g5}\textbf{82.3} \textsubscript{0.2} & \heatcell{g3}13.7 \textsubscript{0.7} & \heatcell{g2}25.4 \textsubscript{0.4} & \heatcell{g2}37.0 \textsubscript{1.0} & \heatcell{g1}47.5 \textsubscript{1.0}

\\ 
& \textsc{25\%} & \heatcell{g1}77.9 \textsubscript{0.0} & \heatcell{g1}76.5 \textsubscript{0.2} & \heatcell{g1}74.8 \textsubscript{0.4} & \heatcell{g1}72.8 \textsubscript{0.3} & \heatcell{g1}71.2 \textsubscript{0.4}
 & \heatcell{g2}81.4 \textsubscript{0.1} & \heatcell{g2}78.7 \textsubscript{0.1} & \heatcell{g3}76.0 \textsubscript{0.3} & \heatcell{g2}72.8 \textsubscript{0.3} & \heatcell{g2}70.2 \textsubscript{0.5}
 & \heatcell{g4}82.1 \textsubscript{0.2} & \heatcell{g5}\textbf{14.0} \textsubscript{0.6} & \heatcell{g3}25.5 \textsubscript{0.8} & \heatcell{g2}37.0 \textsubscript{1.3} & \heatcell{g3}47.7 \textsubscript{0.5}

\\ 
& \textsc{50\%} & \heatcell{g2}78.0 \textsubscript{0.0} & \heatcell{g2}76.6 \textsubscript{0.3} & \heatcell{g1}74.8 \textsubscript{0.4} & \heatcell{g2}73.0 \textsubscript{0.4} & \heatcell{g3}71.6 \textsubscript{0.4}
 & \heatcell{g1}81.2 \textsubscript{0.1} & \heatcell{g1}78.5 \textsubscript{0.2} & \heatcell{g1}75.9 \textsubscript{0.3} & \heatcell{g4}73.1 \textsubscript{0.4} & \heatcell{g4}70.6 \textsubscript{0.5}
 & \heatcell{g1}80.2 \textsubscript{0.3} & \heatcell{g2}13.5 \textsubscript{0.6} & \heatcell{g4}25.6 \textsubscript{0.7} & \heatcell{g2}37.0 \textsubscript{1.4} & \heatcell{g2}47.6 \textsubscript{1.0}

\\ 
& \textsc{75\%} & \heatcell{g2}78.0 \textsubscript{0.0} & \heatcell{g4}76.8 \textsubscript{0.3} & \heatcell{g4}75.1 \textsubscript{0.3} & \heatcell{g4}73.4 \textsubscript{0.4} & \heatcell{g4}71.8 \textsubscript{0.4} 
 & \heatcell{g3}81.5 \textsubscript{0.1} & \heatcell{g4}78.9 \textsubscript{0.3} & \heatcell{g4}76.3 \textsubscript{0.2} & \heatcell{g5}\textbf{73.3} \textsubscript{0.4} & \heatcell{g4}70.6 \textsubscript{0.7}
 & \heatcell{g2}81.6 \textsubscript{0.2} & \heatcell{g4}13.8 \textsubscript{0.4} & \heatcell{g4}25.6 \textsubscript{0.7} & \heatcell{g3}37.3 \textsubscript{1.1} & \heatcell{g4}48.2 \textsubscript{0.7}

\\ 
& \textsc{100\%} & \heatcell{g5}\textbf{78.6} \textsubscript{0.0}  & \heatcell{g5}\textbf{77.3} \textsubscript{0.2}  & \heatcell{g5}\textbf{75.6} \textsubscript{0.3}  & \heatcell{g5}\textbf{74.1} \textsubscript{0.4}  & \heatcell{g5}\textbf{72.8} \textsubscript{0.3}
 & \heatcell{g4}81.7 \textsubscript{0.1} & \heatcell{g5}\textbf{79.0} \textsubscript{0.2} & \heatcell{g5}\textbf{76.4} \textsubscript{0.4} & \heatcell{g5}\textbf{73.3} \textsubscript{0.5} & \heatcell{g5}\textbf{70.8} \textsubscript{0.6}
 & \heatcell{g3}82.0 \textsubscript{0.2} & \heatcell{g3}13.7 \textsubscript{0.4} & \heatcell{g5}\textbf{26.1} \textsubscript{0.6} & \heatcell{g5}\textbf{38.1} \textsubscript{1.8} & \heatcell{g5}\textbf{48.8} \textsubscript{1.2}

\\ 

\bottomrule
\end{tabular}%
}
\caption{Results of evaluating the vanilla non-instruction-tuned baselines (\textsc{van}) and the fine-tuned models under various instruction perturbations using the MMLU, BBH and GSM8K evaluation benchmarks. Results are reported on both the original evaluation instructions (\textsc{0\%}) and the various perturbed evaluation instructions with standard deviations over three runs. \textbf{Bold} values denote the best performance across each model.} 
\label{table:Mix_updated_combined_results}
\end{center}
\end{table*}


\subsection{Evaluation}

\paragraph{General Benchmarks.}
We assess downstream performance using:

\paragraph{Massive Multitask Language Understanding (MMLU; \citealt{hendrycks2021measuring}):} MMLU evaluates a model's factual knowledge and reasoning across 57 subjects, ranging from elementary to professional-level difficulty, using multiple-choice questions.  We follow the original MMLU setup, evaluating in 0-shot and 5-shot settings, and report average test accuracy.

\paragraph{Big-Bench Hard (BBH; \citealt{Suzgun2022ChallengingBT}):} A challenging subset of 23 tasks from the original BIG-Bench \citep{srivastava2022beyond}, aimed at evaluating advanced reasoning in language models. We assess both direct prompting and chain-of-thought (CoT) \cite{10.5555/3600270.3602070}, using official prompts with three in-context examples, and report average exact match across sub-tasks.

\paragraph{Grade School Math (GSM8K; \citealt{Cobbe2021TrainingVT}):} A benchmark of 8.5K  grade school-level word problems for testing multi-step mathematical reasoning in language models. We evaluate with direct prompting and CoT using eight in-context few-shot examples, and we report the exact match.

We follow the same approach as in fine-tuning and create five evaluation instruction sets with different perturbation settings: original instructions (0\%), 25\% of the instructions are perturbed, 50\% are perturbed, 75\% are perturbed, and all instructions are perturbed (100\%). The perturbed instructions are evenly distributed across the six different approaches (see Section~\ref{perturbation_methodolgies}). 

\paragraph{Safety and Bias.}
Additionally, to analyze potential side effects of instruction-tuning on noisy instructions, such as changes in toxicity or misinformation generation, we evaluate the models on: (1) {\bf ToxiGen} \citep{hartvigsen2022toxigen} which measures the extent to which models generate toxic language and hate speech when explicitly prompted to do so across various demographic groups. We report the percentage of toxic outputs identified using a RoBERTa model \citep{liu2020roberta} fine-tuned for toxicity detection, as described by \citet{hartvigsen2022toxigen}; and (2) {\bf TruthfulQA} proposed by \citet{lin2022truthfulqa} which assesses how effectively models can refrain from generating known falsehoods caused by misconceptions or false beliefs, while still generating informative and useful content. We use two off-the-shelf task-specific judge models developed by AllenAI based on Llama-2 (7B) \citep{touvron2023llama} for measuring truthfulness\footnote{\url{https://huggingface.co/allenai/truthfulqa-truth-judge-llama2-7B}} and informativeness, following the setup of \citet{Groeneveld2024OLMoAT}. In both datasets, we evaluate using only the original, unaltered prompts to measure model toxicity and truthfulness.

\section{Results}
\label{sec:results}
Table \ref{table:Mix_updated_combined_results} presents the performance across three benchmarks: MMLU (5-shot), BBH (CoT) and GSM8K (CoT) for each of our models. Full suite of results including 0-shot and direct prompting are presented in  Appendix~\ref{sec:appendix_full_mmlu_results}.

\paragraph{Fine-tuning on perturbed instructions may enhance robustness under noisy prompts.} We first observe that incorporating instruction perturbation during fine-tuning often appears to enhance model robustness across the three evaluation benchmarks and the various evaluation settings. For instance, the Qwen-7B model fine-tuned with 50\% perturbed instructions achieves 0.5\% higher accuracy on MMLU compared to its vanilla (VAN) counterpart when evaluated using 75\% perturbed instructions. In the BBH (CoT) benchmark, it is notable that the Llama-70B model fine-tuned with 100\% perturbed instructions achieves the highest scores when evaluated using 25\%, 50\%, 75\% and 100\% perturbed instructions. However, there are exceptions where the model fine-tuned with the original unaltered instructions still achieves slightly better performance. For example, in the BBH (CoT), the Qwen-72B model fine-tuned on the original unaltered instructions achieves the best overall performance when evaluated using 25\% perturbed instructions. 

\paragraph{Higher proportions of perturbed instructions appear to be beneficial in some contexts.}The results also surprisingly suggest that using a larger number of perturbed instructions in the training mix can lead to improved performance. For example, in both MMLU (5-shot) and BBH (CoT), Qwen-7B, Llama-8B and Llama-70B models achieve their peak performance when fine-tuned with 50\% or more noisy instructions. However, for GSM8K (CoT), smaller models such as Qwen-7B and Llama-8B appear to respond more favorably to less perturbed instructions. One possible explanation for this is that the GSM8K benchmark evaluates multi-step mathematical reasoning, where incomplete or ambiguous instructions can be particularly challenging and harmful for smaller models.

\paragraph{Observed gains on standard unperturbed benchmarks from noisy fine-tuning.} Moreover, the results across all the three benchmarks suggest that fine-tuning on perturbed instructions not only can improve a model's performance under perturbed test conditions but also sometimes yields gains when evaluated on the original, unaltered instructions. For example, in the MMLU (5-shot), when evaluating using the original unaltered instructions, both Llama-8B and Llama-70B models fine-tuned with 100\% perturbed instructions achieve their best performance of 66.0\% and 78.6\% respectively. Similarly, the Qwen-7B model fine-tuned with 75\% perturbed instructions achieves a 0.6\% higher performance than the model variant fine-tuned on the original unaltered instructions when evaluated on the BBH (CoT) benchmark.

\paragraph{CoT remains more effective than direct prompting.}
Prior work has shown that CoT prompting outperforms direct prompting on benchmarks like BBH and GSM8K \cite{NEURIPS2022_8bb0d291,10.5555/3600270.3602070}, and our results confirm this, especially under instruction perturbation. On BBH (CoT), models like Llama-70B fine-tuned with fully perturbed instructions achieve the highest scores of 79.0\% and 76.4\% when evaluated with 25\% and 50\% perturbed instructions, respectively, indicating that CoT prompting benefits from increased robustness introduced during training. While Qwen-72B performs best when fine-tuned with less perturbed instructions, its relatively weaker performance under direct prompting suggests that incorporating perturbation during fine-tuning still contributes to improved generalization and reasoning robustness.

\paragraph{Instruction-tuning yields uneven gains across tasks.}
We observe that the impact of instruction-tuning seems to vary by task. For example, MMLU shows relatively modest improvement from instruction-tuning, regardless of whether or not perturbations are applied. In contrast, BBH appears to consistently benefit from instruction-tuning, especially for Llama models. This observation aligns with findings from \citet{sun-dredze-2025-amuro}, who suggest that certain tasks are already well-represented in a model’s pre-training data, leaving limited room for further gains through instruction-tuning. On the other hand, tasks that are underrepresented or poorly learned during pre-training can potentially see more substantial improvements as the model acquires new task-specific capabilities during instruction-tuning.

\begin{figure}[!t]
    \centering
    \begin{subfigure}[b]{0.5\textwidth}
        \centering
        \includegraphics[width=0.93\linewidth]{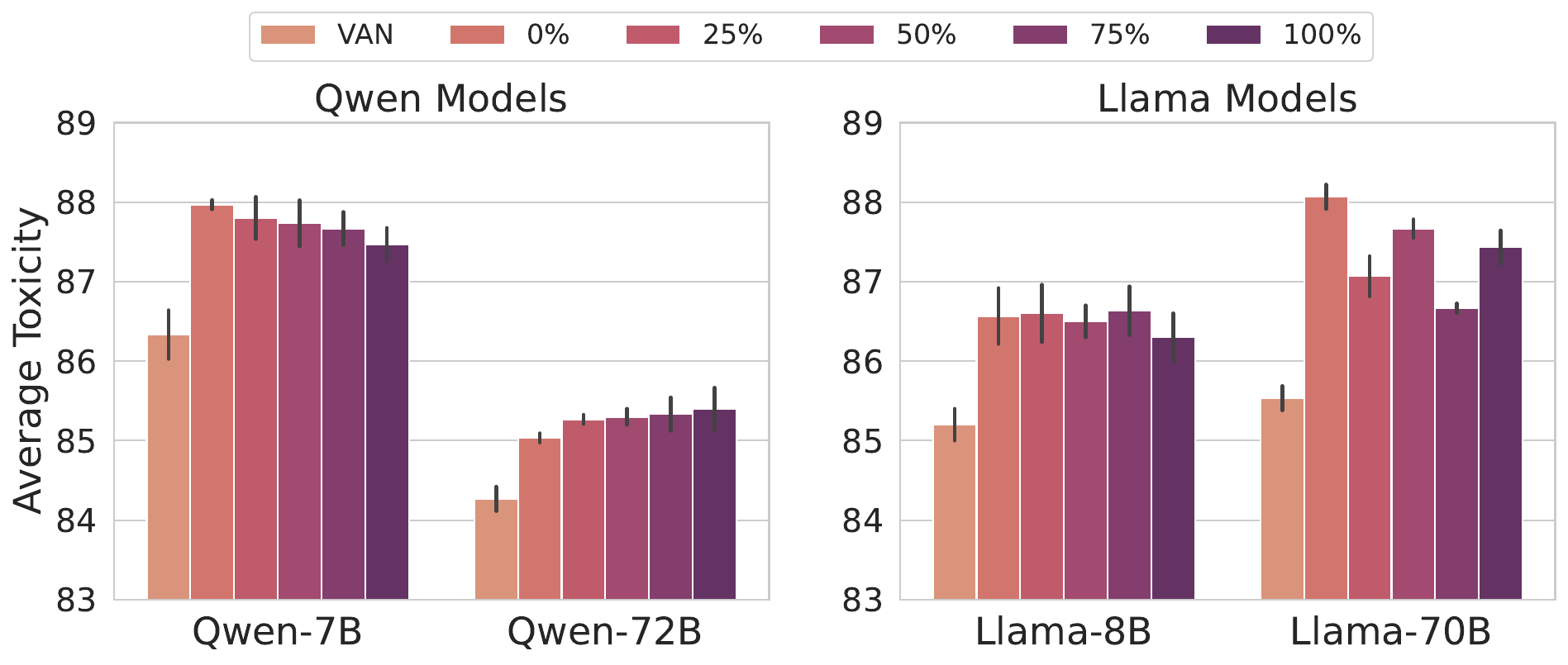}
        \caption{ToxiGen}
        \label{fig:toxigen}
    \end{subfigure}

    \begin{subfigure}[b]{0.5\textwidth}
    \centering
        \includegraphics[width=0.93\linewidth]{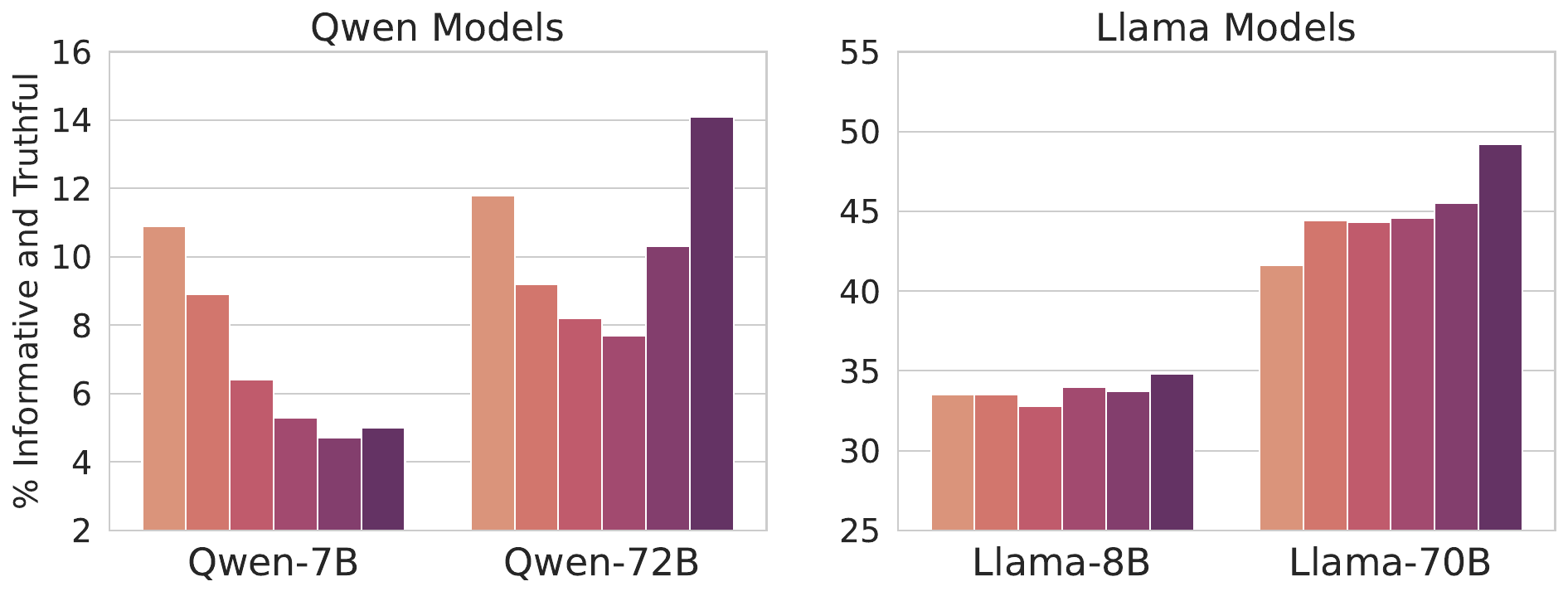}
        \caption{TruthfulQA}
        \label{fig:truthfulqa}
    \end{subfigure}

    \caption{Vanilla non-instruction-tuned baselines (VAN) and the fine-tuned models under various instruction perturbations on ToxiGen benchmark (a) and TruthfulQA (b). Lower is better for toxicity, while higher is better for informative and truthful.}
\label{fig:toxigen_truthfulqa}
\end{figure}

\section{Analysis}


\subsection{Safety and Bias}

Figures~\ref{fig:toxigen} and~\ref{fig:truthfulqa} show model toxicity and truthfulness on the \textit{ToxiGen} and \textit{TruthfulQA} benchmarks respectively, under various instruction perturbations. Fine-tuning with perturbed instructions appears to be associated with enhanced safety and truthfulness across most models. On \textit{ToxiGen}, models like Qwen-7B and Llama-8B exhibit lower average toxicity when fine-tuned with 100\% perturbed instructions, while Llama-70B sees improved results with 75\% perturbation. However, Qwen-72B performs better with original instructions, which may indicate higher sensitivity to perturbations. 

Similarly, on \textit{TruthfulQA}, three out of four models, including Llama-70B, achieve higher truthfulness and informativeness when fine-tuned on fully perturbed data. A possible interpretation is that noisy instructions encourage LLMs to rely less on surface-level patterns and more on robust reasoning. An exception is Qwen-7B, where the vanilla model outperforms all fine-tuned variants, suggesting model-specific sensitivity to instruction noise.

\begin{table}[!t]
\begin{center}
\small
\resizebox{\columnwidth}{!}{%
\begin{tabular}{clc|c|c|c|c}
\toprule
& & {\textbf{MMLU}} & {\textbf{BBH}} & {\textbf{GSM8K}} & {\textbf{TruthfulQA}}  & {\textbf{ToxiGen}} \\

 \textbf{} & 
\textbf{IT} & 5-shot & CoT & CoT & \% Info+True ($\uparrow$) & \% Toxic ($\downarrow$) \\

\midrule

 & 
\textsc{van} & 65.8 \textsubscript{0.0} & 65.8 \textsubscript{1.5} & 55.8 \textsubscript{2.3} & 33.5 \textsubscript{0.0} & 85.4 \textsubscript{0.1}

 \\ 
 \midrule
\multirow{3}{*}{\rotatebox{90}{\footnotesize Dolly}} & \textsc{orig} & 64.2 \textsubscript{0.0} & 62.5 \textsubscript{0.8} & 50.0 \textsubscript{1.6} & 32.8 \textsubscript{0.0} & 87.4 \textsubscript{0.1}
 \\ 
& \textsc{stop} & \textbf{64.6} \textsubscript{0.0} & \textbf{63.1} \textsubscript{1.1} & \textbf{50.5} \textsubscript{2.1} & 35.0 \textsubscript{0.0} & 85.2 \textsubscript{0.3}
 \\ 
& \textsc{shfl} & 64.3 \textsubscript{0.0} & 60.6 \textsubscript{0.9} & \textbf{50.5} \textsubscript{1.8} & \textbf{36.2} \textsubscript{0.0} & \textbf{84.0} \textsubscript{0.1}
 \\ 
 \midrule
\multirow{5}{*}{\rotatebox{90}{\footnotesize GPT4-Alpaca}} &  \textsc{orig} & 64.4 \textsubscript{0.0} & 60.9 \textsubscript{1.3} & 56.0 \textsubscript{2.0} & 60.2 \textsubscript{0.0} & 91.1 \textsubscript{0.1}
 \\ 
& \textsc{shfl} 25\% & 64.5 \textsubscript{0.0} & 61.0 \textsubscript{0.9} & 59.5 \textsubscript{2.3} & 59.5 \textsubscript{0.0} & 90.6 \textsubscript{0.3}
 \\ 
& \textsc{shfl} 50\% & \textbf{64.8} \textsubscript{0.0} & 60.5 \textsubscript{1.3} & \textbf{60.5} \textsubscript{1.9} & 58.5 \textsubscript{0.0} & 90.1 \textsubscript{0.1}
 \\ 
& \textsc{shfl} 75\% & \textbf{64.8} \textsubscript{0.0} & \textbf{62.4} \textsubscript{1.5} & 56.5 \textsubscript{2.1} & 58.9 \textsubscript{0.0} & \textbf{90.0} \textsubscript{0.1}
 \\ 
& \textsc{shfl} 100\% & \textbf{64.8} \textsubscript{0.0} & 62.3 \textsubscript{1.1} & 58.0 \textsubscript{2.0} & \textbf{61.3} \textsubscript{0.0} & 90.2 \textsubscript{0.2}
 \\

\bottomrule
\end{tabular}%
}
\caption{\textit{Llama-8B} trained on  \textit{Dolly} using a single perturbation strategy, removing stop words (\textit{STOP}) or shuffling 25\% of the words (\textit{SHFL}) (top). The same model trained on \textit{GPT4-Alpaca} under varying levels of word shuffling in the instructions (bottom). All instructions were perturbed in each case.} 
\label{table:Ablation_results}
\end{center}
\end{table}

\subsection{Individual Perturbation Strategies}
To better understand the impact of specific perturbation strategies on model performance, we conduct an ablation study using two methods: removing stop words (\textsc{STOP}) and shuffling 25\% of the words in each instruction (\textsc{SHFL}). We fine-tune Llama-8B on Dolly, applying each perturbation strategy to all instructions in the dataset. As a baseline, we also fine-tune the model on the original, unmodified instructions. We evaluate the fine-tuned models on the original, unaltered evaluation benchmarks.

The results in Table~\ref{table:Ablation_results} are broadly consistent with the patterns observed in our main findings (Section~\ref{sec:results}). Notably, the model trained on \textsc{STOP}-perturbed instructions showed improved performance over the one trained on the original dataset across several benchmarks, including MMLU (5-shot), BBH (CoT), and GSM8K (CoT). Interestingly, the \textsc{SHFL} strategy also yields encouraging results. Despite the disruption introduced by shuffling 25\% of the words, the model achieves the highest scores on both the TruthfulQA and ToxiGen benchmarks. These findings suggest that the model may be able to infer the intended task largely from the input data itself, even when the instructions are perturbed. While we initially hypothesized that introducing noise, through stop-word removal or partial word shuffling, would hinder performance, the results indicate a surprising degree of robustness. In some cases, performance even appears to improve under perturbation, suggesting that the model may not rely as heavily on surface-level instruction cues.

\begin{table*}[!t]
  \centering
  \small
  \resizebox{0.9\textwidth}{!}{%

    \begin{tabularx}{0.97\textwidth}{lXcc}
    \toprule
\multicolumn{1}{l}{\textbf{Perturbation}} & \multicolumn{1}{c}{\textbf{Input}} & \multicolumn{1}{c}{\textbf{True}} & \multicolumn{1}{c}{\textbf{Pred}} \\
\hline
    \multirow{2}{*}{(1) Del. Stop.} & \texttt{<instruction>} sentence correct adjective order: \texttt{<\textbackslash instruction>}
    \newline Options: (A) fiberglass old surfboard (B) old fiberglass surfboard & \multirow{2}{*}{(B)} & \multirow{2}{*}{(B) \cmark} \\

    \hline
    
    \multirow{3}{*}{(2) Repl. Wor.} & \texttt{<instruction>} Which sentence has the following adjective order: \texttt{<\textbackslash instruction>}
    \newline Options: (A) driving blue car (B) blue driving car & \multirow{3}{*}{(B)} & \multirow{3}{*}{(A) \xmark} \\
    
    \hline
    
     \multirow{3}{*}{(3) Add Missp.} & \texttt{<instruction>} Which sntnc has the correct adjectivee order: \texttt{<\textbackslash instruction>}
    \newline Options: (A) lovely midsize green Filipino sock (B) Filipino midsize lovely green sock & \multirow{3}{*}{(A)} & \multirow{3}{*}{(A) \cmark} \\
    
    \hline

   \multirow{3}{*}{(4) Del. Wor.} & \texttt{<instruction>} Which has the correct adjective: \texttt{<\textbackslash instruction>}
    \newline Options: (A) plastic grey old-fashioned small sock (B) small old-fashioned grey plastic sock & \multirow{3}{*}{(B)} & \multirow{3}{*}{(A) \xmark} \\
    
    \bottomrule
    \end{tabularx}%
}
\caption{Generated answers by Llama-70B fine-tuned with 100\% perturbed instructions for the \textit{"Which sentence has the correct adjective order"} perturbed question from the BBH benchmark.}
\label{tab:error-examples}
\end{table*}


\subsection{Perturbation Intensity Ablation}
To investigate how the degree of instruction degradation affects model performance, we perform an ablation study that systematically varies the intensity of a word-shuffling perturbation. Specifically, we fine-tune Llama-8B on GPT4-Alpaca with instructions in which 25\%, 50\%, 75\%, and 100\% of the words are randomly shuffled. As a control, we also fine-tune the model on the original, unmodified instructions. All models are then evaluated on the same set of original, unperturbed benchmarks.

The results in Table~\ref{table:Ablation_results} suggest a counterintuitive yet noteworthy trend: as the intensity of perturbation increases, we sometimes observe an improvement in the model's performance. Fine-tuning on instructions with 50\% or more of the words shuffled often yields strong results across all benchmarks. In some cases, the model trained on fully shuffled instructions, where no coherent phrasing remains, performs better than the model trained on the original unperturbed instructions.
These results are broadly consistent with the broader pattern we observed in our main experiments and in the earlier ablation of individual perturbation strategies: performance can improve as superficial instruction cues are degraded. It is possible that heavy instruction noise nudges the model toward the core semantics of the task, which may reduce its dependence on any particular wording and discourage overfitting to fixed prompt templates.

\subsection{Qualitative Analysis}
Table~\ref{tab:error-examples} presents example responses from the Llama-70B model, fine-tuned with fully perturbed instructions, for a sample question from the BBH benchmark. We observe that the model can often produce correct answers even when instructions are altered by removing stop words or introducing misspellings as in examples (1) and (3). However, its performance appears to deteriorate when key words in the instruction are replaced or deleted. For instance, substituting the word ``correct'' with ``following'' as in example (2), or deleting the words ``sentence'' and ``order'' as in example (4), seems to hinder the model's ability to respond correctly.

\section{Theoretical Grounding of Fine-Tuning on Noisy Instructions}
Our findings suggest that incorporating perturbations during instruction-tuning may not only enhance model robustness to noisy or perturbed inputs but may also yield improvements on standard, unperturbed instructions. A plausible explanation for this effect is that noisy instruction-tuning serves as an implicit form of regularization~\cite{bishop1995training}, potentially encouraging models to move beyond reliance on superficial linguistic patterns. Exposure to a wide spectrum of instruction formulations, including those containing syntactic or semantic anomalies, may discourage overfitting to narrow or canonical phrasing. These perturbations effectively broaden the training distribution, functioning as a form of data augmentation~\cite{pmlr-v97-dao19b,hernandez2018data,vaibhav-etal-2019-improving}, and may thereby help LLMs to learn more robust and generalizable task representations.
A particularly noteworthy observation is that models fine-tuned with 100\% perturbed instructions often achieve high accuracy, even when evaluated on standard instructions.
This may suggest that the perturbations could act not merely as noise, but as a potential source of useful inductive bias that enhances generalization across prompt formats.

However, the effectiveness of instruction perturbation appears not to be uniform across all models. While larger models like Llama (70B) and Qwen (72B) exhibit substantial benefits, smaller models, such as Llama (8B) and Qwen (7B), show inconsistent gains. These variations underscore the importance of model-specific calibration of perturbation levels. There may be an upper limit beyond which perturbations become detrimental, particularly for models with limited capacity or for tasks requiring precise instruction-following behavior.

\section{Conclusion}
We explored the impact of instruction perturbation on the robustness and generalization capabilities of instruction-tuned LLMs. By systematically evaluating models of varying sizes across diverse benchmarks, our findings suggest that fine-tuning on structurally perturbed instructions can enhance model performance, particularly under noisy evaluation conditions. Our results indicate that models trained on highly perturbed instructions tend to perform better not only under noisy test conditions but also with standard prompts, suggesting that instruction perturbation encourages more flexible task representations.

These findings point to instruction perturbation as a simple yet potentially effective strategy for enhancing model resilience, particularly in real-world scenarios where user instructions may be inconsistent or ambiguous. By questioning the assumption that clean instructions are always optimal for tuning, this work offers a practical step toward improving instruction-following reliability in large language models. Future research could explore adaptive or semantically-aware perturbation techniques. Such direction may help refine instruction-tuning practices.

\section*{Limitations}

Our experiments were conducted solely with English instructions and downstream tasks due to wide availability of diverse and publicly available instruction-tuning data.  We acknowledge that languages differ in their sensitivity to word order and stop words, a factor not explored in the current work. Chinese, for example, has fewer stop words and a less rigid syntactic structure than English, allowing for greater flexibility in word order.  Therefore, the effects of perturbation should be investigated with respect to the specific linguistic characteristics of each language under consideration in future work.

\section*{Acknowledgments}
We thank Atsuki Yamaguchi and the
anonymous reviewers for their invaluable feedback.

AA is supported by the Centre for Doctoral Training in Speech and Language Technologies (SLT) and their Applications funded by UK Research and Innovation grant EP/S023062/1. XT and NA are supported by the EPSRC [grant number EP/Y009800/1], through funding from Responsible AI UK (KP0016) as a Keystone project.
We acknowledge the IT Services at the University of Sheffield for Supercomputing for the provision of HPC services.

\bibliography{custom}

\appendix
\clearpage 

\section*{Appendix}

\section{Instruction-tuning Hyperparameters}
\label{sec:appendix_hyperparameters}
For instruction-tuning and evaluation, we adopt the implementation from Open Instruct~\citep{wang2023far,ivison2023camels,ivison2024unpacking,Lambert2024TLU3P}. Table \ref{table:param_instruction_tuning} shows the hyperparameters used in our instruction fine-tuning experiments.

\renewcommand*{\arraystretch}{1.1}
\begin{table}[ht]
\begin{center}
\small
\resizebox{0.3\textwidth}{!}{%
\begin{tabular}{lc}
\toprule
\textbf{Hyperparameter} & \textbf{Value}  \\ \midrule
learning rate & 1e-5 \\
lr scheduler type & linear \\
warmup ratio & 0.03 \\
weight decay & 0 \\
\# train epochs & 1 \\
gradient acc. steps & 128 \\
max. seq. length & 4,096 \\
temperature & 0.01 \\
LoRA rank & 64 \\
LoRA alpha & 16 \\
LoRA dropout & 0.1 \\
\bottomrule
\end{tabular}%
}
\caption{The instruction fine-tuning hyperparameters.} 
\label{table:param_instruction_tuning}
\end{center}
\end{table}


\section{Full Results}
\label{sec:appendix_full_mmlu_results}
In addition to the results on MMLU (5-shot), BBH (CoT) and GSM8K (CoT) presented in Table~\ref{table:Mix_updated_combined_results}, we present the full results on MMLU (Table~\ref{table:Mix_updated_mmlu_results}), BBH (Table~\ref{table:Mix_updated_bbh_results}) and GSM8K (Table~\ref{table:Mix_updated_GSM_results}).

\begin{table*}[!t]
\begin{center}
\small
\resizebox{0.83\textwidth}{!}{%
\begin{tabular}{clccccc|ccccc}
\toprule
& & \multicolumn{5}{c}{\textbf{MMLU (0-shot)}} & \multicolumn{5}{c}{\textbf{MMLU (5-shot)}} \\

\textbf{} & \textbf{IT} & \textsc{0\%} & \textsc{25\%} & \textsc{50\%} & \textsc{75\%} & \textsc{100\%} & \textsc{0\%} & \textsc{25\%} & \textsc{50\%} & \textsc{75\%} & \textsc{100\%}\\

\midrule

\multirow{6}{*}{\rotatebox{90}{\footnotesize Qwen 7B}}
& \textsc{van.} & 72.0 \textsubscript{0.0} & 70.6 \textsubscript{0.0} & 69.2 \textsubscript{0.3} & 67.8 \textsubscript{0.6} & 66.3 \textsubscript{0.4} & 74.3 \textsubscript{0.0} & 73.0 \textsubscript{0.1} & 71.5 \textsubscript{0.1} & 70.0 \textsubscript{0.4} & 68.6 \textsubscript{0.6}
 \\ 
 \cmidrule{2-12}
& \textsc{0\%} & \heatcell{g4}72.3 \textsubscript{0.0} & \heatcell{g1}71.0 \textsubscript{0.0} & \heatcell{g1}69.8 \textsubscript{0.1} & \heatcell{g2}68.4 \textsubscript{0.6} & \heatcell{g1}66.9 \textsubscript{0.4} & \heatcell{g1}74.3 \textsubscript{0.0} & \heatcell{g1}73.0 \textsubscript{0.1} & \heatcell{g1}71.7 \textsubscript{0.1} & \heatcell{g1}70.2 \textsubscript{0.4} & \heatcell{g1}68.9 \textsubscript{0.7}
 \\ 
& \textsc{25\%} & \heatcell{g4}72.3 \textsubscript{0.0} & \heatcell{g5}\textbf{71.1} \textsubscript{0.1} & \heatcell{g4}69.9 \textsubscript{0.2} & \heatcell{g1}68.3 \textsubscript{0.5} & \heatcell{g2}67.1 \textsubscript{0.4} & \heatcell{g5}\textbf{74.4} \textsubscript{0.0} & \heatcell{g1}73.0 \textsubscript{0.1} & \heatcell{g2}71.8 \textsubscript{0.1} & \heatcell{g2}70.3 \textsubscript{0.5} & \heatcell{g2}69.1 \textsubscript{0.6}
 \\ 
& \textsc{50\%} & \heatcell{g1}72.2 \textsubscript{0.0} & \heatcell{g1}71.0 \textsubscript{0.1} & \heatcell{g1}69.8 \textsubscript{0.2} & \heatcell{g2}68.4 \textsubscript{0.6} & \heatcell{g2}67.1 \textsubscript{0.5} & \heatcell{g5}\textbf{74.4} \textsubscript{0.0} & \heatcell{g5}\textbf{73.1} \textsubscript{0.1} & \heatcell{g5}\textbf{71.9} \textsubscript{0.1} & \heatcell{g5}\textbf{70.5} \textsubscript{0.5} & \heatcell{g2}69.1 \textsubscript{0.7}
 \\ 
& \textsc{75\%} & \heatcell{g5}\textbf{72.4} \textsubscript{0.0} & \heatcell{g5}\textbf{71.1} \textsubscript{0.2} & \heatcell{g5}\textbf{70.0} \textsubscript{0.2} & \heatcell{g4}68.5 \textsubscript{0.6} & \heatcell{g4}67.3 \textsubscript{0.3} & \heatcell{g1}74.3 \textsubscript{0.0} & \heatcell{g1}73.0 \textsubscript{0.1} & \heatcell{g2}71.8 \textsubscript{0.2} & \heatcell{g4}70.4 \textsubscript{0.5} & \heatcell{g2}69.1 \textsubscript{0.7}
 \\ 
& \textsc{100\%} & \heatcell{g5}\textbf{72.4} \textsubscript{0.0} & \heatcell{g5}\textbf{71.1} \textsubscript{0.1} & \heatcell{g5}\textbf{70.0} \textsubscript{0.2} & \heatcell{g5}\textbf{68.6} \textsubscript{0.5} & \heatcell{g5}\textbf{67.4} \textsubscript{0.3} & \heatcell{g1}74.3 \textsubscript{0.0} & \heatcell{g5}\textbf{73.1} \textsubscript{0.0} & \heatcell{g5}\textbf{71.9} \textsubscript{0.1} & \heatcell{g5}\textbf{70.5} \textsubscript{0.5} & \heatcell{g5}\textbf{69.2} \textsubscript{0.6}
 \\ 

\midrule

\multirow{6}{*}{\rotatebox{90}{\footnotesize Llama 8B}}
& \textsc{van.} & 64.2 \textsubscript{0.0} & 62.6 \textsubscript{0.1} & 60.9 \textsubscript{0.2} & 59.4 \textsubscript{0.0} & 57.5 \textsubscript{0.2} & 65.8 \textsubscript{0.0} & 64.5 \textsubscript{0.1} & 63.1 \textsubscript{0.1} & 62.1 \textsubscript{0.3} & 60.8 \textsubscript{0.5}
 \\ 
 \cmidrule{2-12}
& \textsc{0\%} & \heatcell{g5}\textbf{65.1} \textsubscript{0.0}  & \heatcell{g5}\textbf{63.6} \textsubscript{0.2}  & \heatcell{g4}61.9 \textsubscript{0.4} & \heatcell{g3}60.3 \textsubscript{0.1} & \heatcell{g2}58.5 \textsubscript{0.3} & \heatcell{g2}65.8 \textsubscript{0.0} & \heatcell{g1}64.6 \textsubscript{0.2} & \heatcell{g1}63.3 \textsubscript{0.1} & \heatcell{g1}62.2 \textsubscript{0.2} & \heatcell{g1}60.7 \textsubscript{0.5}
 \\ 
& \textsc{25\%} & \heatcell{g2}64.4 \textsubscript{0.0} & \heatcell{g2}63.0 \textsubscript{0.2} & \heatcell{g1}61.5 \textsubscript{0.4} & \heatcell{g1}60.0 \textsubscript{0.3} & \heatcell{g1}58.4 \textsubscript{0.4} & \heatcell{g4}65.9 \textsubscript{0.0} & \heatcell{g5}\textbf{64.8} \textsubscript{0.3}  & \heatcell{g2}63.4 \textsubscript{0.2} & \heatcell{g2}62.3 \textsubscript{0.2} & \heatcell{g2}60.9 \textsubscript{0.7}
 \\ 
& \textsc{50\%} & \heatcell{g1}64.3 \textsubscript{0.0} & \heatcell{g1}62.9 \textsubscript{0.2} & \heatcell{g2}61.6 \textsubscript{0.4} & \heatcell{g2}60.2 \textsubscript{0.2} & \heatcell{g3}58.6 \textsubscript{0.3} & \heatcell{g4}65.9 \textsubscript{0.0} & \heatcell{g5}\textbf{64.8} \textsubscript{0.2}  & \heatcell{g4}63.6 \textsubscript{0.1} & \heatcell{g5}\textbf{62.5} \textsubscript{0.2}  & \heatcell{g4}61.0 \textsubscript{0.6}
 \\ 
& \textsc{75\%} & \heatcell{g4}64.9 \textsubscript{0.0} & \heatcell{g4}63.5 \textsubscript{0.2} & \heatcell{g5}\textbf{62.0} \textsubscript{0.5}  & \heatcell{g4}60.5 \textsubscript{0.1} & \heatcell{g5}\textbf{58.8} \textsubscript{0.4}  & \heatcell{g1}65.7 \textsubscript{0.0} & \heatcell{g4}64.7 \textsubscript{0.2} & \heatcell{g4}63.6 \textsubscript{0.1} & \heatcell{g5}\textbf{62.5} \textsubscript{0.2}  & \heatcell{g5}\textbf{61.2} \textsubscript{0.5} 
 \\ 
& \textsc{100\% } & \heatcell{g3}64.7 \textsubscript{0.0} & \heatcell{g3}63.4 \textsubscript{0.2} & \heatcell{g4}61.9 \textsubscript{0.5} & \heatcell{g5}\textbf{60.6} \textsubscript{0.1}  & \heatcell{g4}58.7 \textsubscript{0.3} & \heatcell{g5}\textbf{66.0} \textsubscript{0.0}  & \heatcell{g5}\textbf{64.8} \textsubscript{0.3}  & \heatcell{g5}\textbf{63.7} \textsubscript{0.1}  & \heatcell{g5}\textbf{62.5} \textsubscript{0.3}  & \heatcell{g5}\textbf{61.2} \textsubscript{0.5} 
 \\ 

\midrule

\multirow{6}{*}{\rotatebox{90}{\footnotesize Qwen 72B}}
& \textsc{van.} & 83.1 \textsubscript{0.0} & 81.7 \textsubscript{0.2} & 80.5 \textsubscript{0.2} & 78.9 \textsubscript{0.3} & 77.3 \textsubscript{0.3} & 85.7 \textsubscript{0.0} & 84.5 \textsubscript{0.2} & 83.0 \textsubscript{0.3} & 81.8 \textsubscript{0.3} & 80.3 \textsubscript{0.4}
 \\ 
 \cmidrule{2-12}
& \textsc{0\%} & \heatcell{g5}\textbf{83.8} \textsubscript{0.0}  & \heatcell{g5}\textbf{82.4} \textsubscript{0.1}  & \heatcell{g1}81.1 \textsubscript{0.2} & \heatcell{g1}79.5 \textsubscript{0.3} & \heatcell{g1}77.9 \textsubscript{0.3} & \heatcell{g5}\textbf{85.8} \textsubscript{0.0}  & \heatcell{g1}84.6 \textsubscript{0.2} & \heatcell{g1}83.1 \textsubscript{0.3} & \heatcell{g1}82.0 \textsubscript{0.2} & \heatcell{g1}80.5 \textsubscript{0.5} \\ 
& \textsc{25\%} & \heatcell{g2}83.7 \textsubscript{0.0} & \heatcell{g1}82.3 \textsubscript{0.1} & \heatcell{g1}81.1 \textsubscript{0.3} & \heatcell{g2}79.6 \textsubscript{0.4} & \heatcell{g2}78.0 \textsubscript{0.2} & \heatcell{g1}85.7 \textsubscript{0.0} & \heatcell{g1}84.6 \textsubscript{0.3} & \heatcell{g1}83.1 \textsubscript{0.3} & \heatcell{g1}82.0 \textsubscript{0.3} & \heatcell{g1}80.5 \textsubscript{0.6} \\ 
& \textsc{50\%} & \heatcell{g1}83.6 \textsubscript{0.0} & \heatcell{g1}82.3 \textsubscript{0.1} & \heatcell{g1}81.1 \textsubscript{0.3} & \heatcell{g3}79.7 \textsubscript{0.3} & \heatcell{g4}78.2 \textsubscript{0.2} & \heatcell{g1}85.7 \textsubscript{0.0} & \heatcell{g2}84.7 \textsubscript{0.3} & \heatcell{g1}83.1 \textsubscript{0.3} & \heatcell{g1}82.0 \textsubscript{0.3} & \heatcell{g5}\textbf{80.7} \textsubscript{0.6}  \\ 
& \textsc{75\%} & \heatcell{g2}83.7 \textsubscript{0.0} & \heatcell{g5}\textbf{82.4} \textsubscript{0.2}  & \heatcell{g5}\textbf{81.2} \textsubscript{0.2}  & \heatcell{g4}79.9 \textsubscript{0.3} & \heatcell{g5}\textbf{78.3} \textsubscript{0.2}  & \heatcell{g5}\textbf{85.8} \textsubscript{0.0}  & \heatcell{g2}84.7 \textsubscript{0.3} & \heatcell{g5}\textbf{83.2} \textsubscript{0.3}  & \heatcell{g5}\textbf{82.1} \textsubscript{0.3}  & \heatcell{g5}\textbf{80.7} \textsubscript{0.5}  \\ 
& \textsc{100\%} & \heatcell{g5}\textbf{83.8} \textsubscript{0.0}  & \heatcell{g5}\textbf{82.4} \textsubscript{0.1}  & \heatcell{g5}\textbf{81.2} \textsubscript{0.2}  & \heatcell{g5}\textbf{80.0} \textsubscript{0.3}  & \heatcell{g5}\textbf{78.3} \textsubscript{0.2}  & \heatcell{g5}\textbf{85.8} \textsubscript{0.0}  & \heatcell{g5}\textbf{84.8} \textsubscript{0.2}  & \heatcell{g5}\textbf{83.2} \textsubscript{0.3}  & \heatcell{g5}\textbf{82.1} \textsubscript{0.4} & \heatcell{g3}80.6 \textsubscript{0.6} \\ 

\midrule

\multirow{6}{*}{\rotatebox{90}{\footnotesize Llama 70B}}
& \textsc{van.} & 74.4 \textsubscript{0.0} & 72.8 \textsubscript{0.2} & 71.1 \textsubscript{0.1} & 69.0 \textsubscript{0.2} & 67.3 \textsubscript{0.3} & 75.8 \textsubscript{0.0} & 74.1 \textsubscript{0.1} & 72.2 \textsubscript{0.2} & 70.2 \textsubscript{0.4} & 68.5 \textsubscript{0.4}
 \\ 
 \cmidrule{2-12}
& \textsc{0\%} & \heatcell{g4}75.7 \textsubscript{0.0} & \heatcell{g3}74.0 \textsubscript{0.3} & \heatcell{g2}72.6 \textsubscript{0.2} & \heatcell{g2}70.8 \textsubscript{0.1} & \heatcell{g2}69.5 \textsubscript{0.3} & \heatcell{g4}78.1 \textsubscript{0.0} & \heatcell{g3}76.7 \textsubscript{0.3} & \heatcell{g2}74.9 \textsubscript{0.3} & \heatcell{g2}73.0 \textsubscript{0.4} & \heatcell{g2}71.4 \textsubscript{0.5} \\ 
& \textsc{25\%} & \heatcell{g2}75.5 \textsubscript{0.0} & \heatcell{g1}73.8 \textsubscript{0.3} & \heatcell{g1}72.4 \textsubscript{0.3} & \heatcell{g1}70.7 \textsubscript{0.1} & \heatcell{g1}69.2 \textsubscript{0.3} & \heatcell{g1}77.9 \textsubscript{0.0} & \heatcell{g1}76.5 \textsubscript{0.2} & \heatcell{g1}74.8 \textsubscript{0.4} & \heatcell{g1}72.8 \textsubscript{0.3} & \heatcell{g1}71.2 \textsubscript{0.4} \\ 
& \textsc{50\%} & \heatcell{g1}75.4 \textsubscript{0.0} & \heatcell{g2}73.9 \textsubscript{0.3} & \heatcell{g3}72.7 \textsubscript{0.1} & \heatcell{g3}70.9 \textsubscript{0.1} & \heatcell{g3}69.6 \textsubscript{0.3} & \heatcell{g2}78.0 \textsubscript{0.0} & \heatcell{g2}76.6 \textsubscript{0.3} & \heatcell{g1}74.8 \textsubscript{0.4} & \heatcell{g2}73.0 \textsubscript{0.4} & \heatcell{g3}71.6 \textsubscript{0.4} \\ 
& \textsc{75\%} & \heatcell{g3}75.6 \textsubscript{0.0} & \heatcell{g4}74.1 \textsubscript{0.2} & \heatcell{g4}72.8 \textsubscript{0.2} & \heatcell{g4}71.0 \textsubscript{0.1} & \heatcell{g4}69.7 \textsubscript{0.3} & \heatcell{g2}78.0 \textsubscript{0.0} & \heatcell{g4}76.8 \textsubscript{0.3} & \heatcell{g4}75.1 \textsubscript{0.3} & \heatcell{g4}73.4 \textsubscript{0.4} & \heatcell{g4}71.8 \textsubscript{0.4} \\ 
& \textsc{100\%} & \heatcell{g5}\textbf{76.6} \textsubscript{0.0}  & \heatcell{g5}\textbf{75.1} \textsubscript{0.2}  & \heatcell{g5}\textbf{73.7} \textsubscript{0.1}  & \heatcell{g5}\textbf{72.0} \textsubscript{0.0}  & \heatcell{g5}\textbf{70.7} \textsubscript{0.3}  & \heatcell{g5}\textbf{78.6} \textsubscript{0.0}  & \heatcell{g5}\textbf{77.3} \textsubscript{0.2}  & \heatcell{g5}\textbf{75.6} \textsubscript{0.3}  & \heatcell{g5}\textbf{74.1} \textsubscript{0.4}  & \heatcell{g5}\textbf{72.8} \textsubscript{0.3}  \\ 

\bottomrule
\end{tabular}%
}
\caption{Results of evaluating the fine-tuned models under various instruction perturbations using the MMLU evaluation benchmark. Accuracy is reported on both the original evaluation instructions (\textsc{0\%}) and the various perturbed evaluation instructions. \textbf{Bold} values denote the best performance across each model.} 
\label{table:Mix_updated_mmlu_results}
\end{center}
\end{table*}


\begin{table*}[!t]
\begin{center}
\small
\resizebox{0.83\textwidth}{!}{%
\begin{tabular}{clccccc|ccccc}
\toprule
& & \multicolumn{5}{c}{\textbf{BBH (CoT)}} & \multicolumn{5}{c}{\textbf{BBH (direct)}} \\

\textbf{} & \textbf{IT} & \textsc{0\%} & \textsc{25\%} & \textsc{50\%} & \textsc{75\%} & \textsc{100\%} & \textsc{0\%} & \textsc{25\%} & \textsc{50\%} & \textsc{75\%} & \textsc{100\%}\\

\midrule

\multirow{6}{*}{\rotatebox{90}{\footnotesize Qwen 7B}}
& \textsc{van.}  & 66.7 \textsubscript{0.1} & 63.9 \textsubscript{0.4} & 60.8 \textsubscript{0.4} & 57.7 \textsubscript{0.5} & 54.9 \textsubscript{0.5}  & 31.0 \textsubscript{0.1}  & 27.0 \textsubscript{0.1}  & 22.9 \textsubscript{0.4}  & 18.6 \textsubscript{0.2}  & 14.5 \textsubscript{0.4}

 \\ 
 \cmidrule{2-12}
& \textsc{0\%} & \heatcell{g3}66.8 \textsubscript{0.0} & \heatcell{g1}62.7 \textsubscript{0.2} & \heatcell{g1}58.7 \textsubscript{0.2} & \heatcell{g1}54.3 \textsubscript{0.5} & \heatcell{g1}50.6 \textsubscript{0.6}  & \heatcell{g5}\textbf{50.5} \textsubscript{0.0}  & \heatcell{g5}\textbf{48.9} \textsubscript{0.2}  & \heatcell{g5}\textbf{47.0} \textsubscript{0.3}  & \heatcell{g5}\textbf{45.1} \textsubscript{0.6}  & \heatcell{g5}\textbf{43.2} \textsubscript{0.7}

 \\ 
& \textsc{25\%} & \heatcell{g2}66.7 \textsubscript{0.0} & \heatcell{g2}63.3 \textsubscript{0.3} & \heatcell{g2}59.7 \textsubscript{0.2} & \heatcell{g2}55.9 \textsubscript{0.5} & \heatcell{g2}52.4 \textsubscript{0.6}  & \heatcell{g3}50.2 \textsubscript{0.0}  & \heatcell{g3}48.4 \textsubscript{0.1}  & \heatcell{g3}46.5 \textsubscript{0.2}  & \heatcell{g3}44.4 \textsubscript{0.7}  & \heatcell{g4}42.3 \textsubscript{0.7}

 \\ 
& \textsc{50\%} & \heatcell{g4}67.0 \textsubscript{0.0} & \heatcell{g5}\textbf{64.0} \textsubscript{0.2} & \heatcell{g5}\textbf{61.1} \textsubscript{0.3} & \heatcell{g5}\textbf{57.7} \textsubscript{0.6} & \heatcell{g5}\textbf{54.8} \textsubscript{0.6} & \heatcell{g3}50.2 \textsubscript{0.0}  & \heatcell{g2}48.3 \textsubscript{0.2}  & \heatcell{g1}46.3 \textsubscript{0.2}  & \heatcell{g1}43.9 \textsubscript{0.5}  & \heatcell{g2}41.9 \textsubscript{0.7}

 \\ 
& \textsc{75\%} & \heatcell{g5}\textbf{67.4} \textsubscript{0.0} & \heatcell{g4}63.9 \textsubscript{0.3} & \heatcell{g4}60.7 \textsubscript{0.2} & \heatcell{g4}57.6 \textsubscript{0.4} & \heatcell{g4}54.7 \textsubscript{0.6} & \heatcell{g4}50.4 \textsubscript{0.0}  & \heatcell{g4}48.7 \textsubscript{0.2}  & \heatcell{g4}46.9 \textsubscript{0.3}  & \heatcell{g4}44.9 \textsubscript{0.7}  & \heatcell{g5}\textbf{43.2} \textsubscript{0.7}

 \\ 
& \textsc{100\%} & \heatcell{g1}66.6 \textsubscript{0.0} & \heatcell{g3}63.4 \textsubscript{0.2} & \heatcell{g3}60.3 \textsubscript{0.3} & \heatcell{g3}56.8 \textsubscript{0.4} & \heatcell{g3}53.8 \textsubscript{0.7} & \heatcell{g2}50.0 \textsubscript{0.0}  & \heatcell{g1}48.2 \textsubscript{0.2}  & \heatcell{g2}46.4 \textsubscript{0.2}  & \heatcell{g2}44.1 \textsubscript{0.3}  & \heatcell{g3}42.1 \textsubscript{0.5}

 \\ 

\midrule

\multirow{6}{*}{\rotatebox{90}{\footnotesize Llama 8B}} 
& \textsc{van.} & 64.5 \textsubscript{0.1}  & 62.5 \textsubscript{0.3}  & 60.2 \textsubscript{0.4}  & 57.5 \textsubscript{1.0}  & 55.0 \textsubscript{0.9}  & 45.7 \textsubscript{0.1}  & 44.4 \textsubscript{0.3}  & 43.0 \textsubscript{0.2}  & 41.4 \textsubscript{0.3}  & 40.1 \textsubscript{0.7}

 \\ 
 \cmidrule{2-12}
& \textsc{0\%} & \heatcell{g3}63.0 \textsubscript{0.4} & \heatcell{g2}63.4 \textsubscript{0.1} & \heatcell{g2}61.1 \textsubscript{0.3} & \heatcell{g1}58.7 \textsubscript{0.7} & \heatcell{g3}56.5 \textsubscript{0.6}  & \heatcell{g1}45.1 \textsubscript{0.4}  & \heatcell{g4}46.0 \textsubscript{0.3}  & \heatcell{g2}44.2 \textsubscript{0.3}  & \heatcell{g2}42.3 \textsubscript{0.1}  & \heatcell{g1}40.7 \textsubscript{0.5}

 \\ 
& \textsc{25\%} & \heatcell{g4}66.0 \textsubscript{0.1} & \heatcell{g1}60.5 \textsubscript{0.4} & \heatcell{g1}60.0 \textsubscript{1.9} & \heatcell{g3}59.1 \textsubscript{0.4} & \heatcell{g3}56.5 \textsubscript{0.6} & \heatcell{g5}\textbf{47.4} \textsubscript{0.1}  & \heatcell{g2}44.3 \textsubscript{0.3}  & \heatcell{g1}43.6 \textsubscript{0.8} & \heatcell{g4}42.6 \textsubscript{0.3} & \heatcell{g4}41.3 \textsubscript{0.4}

 \\ 
& \textsc{50\%}  & \heatcell{g1}62.7 \textsubscript{0.0} & \heatcell{g5}\textbf{64.4} \textsubscript{0.3} & \heatcell{g5}\textbf{62.0} \textsubscript{0.5} & \heatcell{g5}\textbf{59.3} \textsubscript{0.5} & \heatcell{g4}56.7 \textsubscript{0.5} & \heatcell{g3}46.1 \textsubscript{0.1}  & \heatcell{g5}\textbf{46.3} \textsubscript{0.2}  & \heatcell{g5}\textbf{44.6} \textsubscript{0.2}  & \heatcell{g5}\textbf{42.8} \textsubscript{0.3}  & \heatcell{g5}\textbf{41.5} \textsubscript{0.3}

 \\
& \textsc{75\%}  & \heatcell{g2}62.9 \textsubscript{0.4} & \heatcell{g3}64.1 \textsubscript{0.3} & \heatcell{g3}61.8 \textsubscript{0.3} & \heatcell{g3}59.1 \textsubscript{0.5} & \heatcell{g2}56.3 \textsubscript{0.5} & \heatcell{g2}45.3 \textsubscript{0.5}  & \heatcell{g3}45.8 \textsubscript{0.1}  & \heatcell{g3}44.3 \textsubscript{0.4}  & \heatcell{g3}42.4 \textsubscript{0.1}  & \heatcell{g2}40.9 \textsubscript{0.3}

 \\ 
& \textsc{100\%}  & \heatcell{g5}\textbf{66.2} \textsubscript{0.1} & \heatcell{g4}64.2 \textsubscript{0.5} & \heatcell{g4}62.0 \textsubscript{0.5} & \heatcell{g4}59.2 \textsubscript{0.8} & \heatcell{g5}\textbf{56.8} \textsubscript{0.4} & \heatcell{g4}47.3 \textsubscript{0.1}  & \heatcell{g3}45.8 \textsubscript{0.3}  & \heatcell{g4}44.5 \textsubscript{0.2} & \heatcell{g4}42.6 \textsubscript{0.1}  & \heatcell{g3}41.2 \textsubscript{0.5}

 \\ 

\midrule

\multirow{6}{*}{\rotatebox{90}{\footnotesize Qwen 72B}}
& \textsc{van.} & 82.7 \textsubscript{0.1} & 79.2 \textsubscript{0.2} & 75.4 \textsubscript{0.2}  & 71.7 \textsubscript{0.4}  & 68.1 \textsubscript{0.8}  & 26.4 \textsubscript{0.1}  & 23.8 \textsubscript{0.4}  & 21.2 \textsubscript{0.4}  & 18.4 \textsubscript{0.4}  & 15.8 \textsubscript{0.2}

 \\ 
 \cmidrule{2-12}
& \textsc{0\%} & \heatcell{g5}\textbf{83.8} \textsubscript{0.1} & \heatcell{g5}\textbf{80.5} \textsubscript{0.2} & \heatcell{g4}77.3 \textsubscript{0.3} & \heatcell{g3}73.8 \textsubscript{0.5} & \heatcell{g5}\textbf{70.8} \textsubscript{1.1} & \heatcell{g4}66.6 \textsubscript{0.1}  & \heatcell{g4}64.5 \textsubscript{0.2}  & \heatcell{g3}62.2 \textsubscript{0.2}  & \heatcell{g2}59.7 \textsubscript{0.4}  & \heatcell{g3}57.6 \textsubscript{0.8}
 \\ 
& \textsc{25\%}  & \heatcell{g5}\textbf{83.8} \textsubscript{0.1} & \heatcell{g4}80.4 \textsubscript{0.2} & \heatcell{g5}\textbf{77.4} \textsubscript{0.4} & \heatcell{g5}\textbf{74.0} \textsubscript{0.8} & \heatcell{g5}\textbf{70.8} \textsubscript{1.0}  & \heatcell{g2}66.5 \textsubscript{0.0}  & \heatcell{g4}64.5 \textsubscript{0.2}  & \heatcell{g3}62.2 \textsubscript{0.2}  & \heatcell{g2}59.7 \textsubscript{0.6}  & \heatcell{g2}57.5 \textsubscript{0.8}
 \\ 
& \textsc{50\%}  & \heatcell{g2}83.3 \textsubscript{0.1} & \heatcell{g1}80.2 \textsubscript{0.2} & \heatcell{g2}77.2 \textsubscript{0.2} & \heatcell{g1}73.7 \textsubscript{0.6} & \heatcell{g4}70.6 \textsubscript{0.8} & \heatcell{g2}66.5 \textsubscript{0.1}  & \heatcell{g2}64.4 \textsubscript{0.3}  & \heatcell{g3}62.2 \textsubscript{0.5}  & \heatcell{g3}59.9 \textsubscript{0.6}  & \heatcell{g3}57.6 \textsubscript{0.7}
 \\ 
& \textsc{75\%}  & \heatcell{g4}83.6 \textsubscript{0.0} & \heatcell{g3}80.3 \textsubscript{0.2} & \heatcell{g2}77.2 \textsubscript{0.3} & \heatcell{g4}73.9 \textsubscript{0.6} & \heatcell{g2}70.4 \textsubscript{0.8} & \heatcell{g2}66.5 \textsubscript{0.0}  & \heatcell{g4}64.5 \textsubscript{0.2}  & \heatcell{g3}62.2 \textsubscript{0.5}  & \heatcell{g4}60.0 \textsubscript{0.6}  & \heatcell{g3}57.6 \textsubscript{0.8}
 \\ 
& \textsc{100\%}  & \heatcell{g4}83.6 \textsubscript{0.0} & \heatcell{g4}80.4 \textsubscript{0.2} & \heatcell{g4}77.3 \textsubscript{0.2} & \heatcell{g3}73.8 \textsubscript{0.4} & \heatcell{g5}\textbf{70.8} \textsubscript{0.8}  & \heatcell{g5}\textbf{67.1} \textsubscript{0.0}  & \heatcell{g5}\textbf{65.0} \textsubscript{0.3}  & \heatcell{g5}\textbf{62.7} \textsubscript{0.5}  & \heatcell{g5}\textbf{60.3} \textsubscript{0.4}  & \heatcell{g5}\textbf{58.0} \textsubscript{0.6}
 \\ 

\midrule

\multirow{6}{*}{\rotatebox{90}{\footnotesize Llama 70B}}
& \textsc{van.} & 78.3 \textsubscript{0.1}  & 75.7 \textsubscript{0.2}  & 73.3 \textsubscript{0.2}  & 70.3 \textsubscript{0.4}  & 68.1 \textsubscript{0.4}  & 58.1 \textsubscript{0.1}  & 56.5 \textsubscript{0.3}  & 54.9 \textsubscript{0.5}  & 53.0 \textsubscript{0.8}  & 51.3 \textsubscript{0.8}

 \\ 
 \cmidrule{2-12}
& \textsc{0\%}  & \heatcell{g5}\textbf{81.8} \textsubscript{0.1} & \heatcell{g4}78.9 \textsubscript{0.1} & \heatcell{g1}75.9 \textsubscript{0.2} & \heatcell{g1}72.7 \textsubscript{0.4} & \heatcell{g1}70.1 \textsubscript{0.4}  & \heatcell{g4}63.9 \textsubscript{0.1}  & \heatcell{g4}61.7 \textsubscript{0.2}  & \heatcell{g4}59.2 \textsubscript{0.5}  & \heatcell{g2}56.3 \textsubscript{0.7}  & \heatcell{g1}53.7 \textsubscript{0.4}
 \\ 
& \textsc{25\%}  & \heatcell{g2}81.4 \textsubscript{0.1} & \heatcell{g2}78.7 \textsubscript{0.1} & \heatcell{g3}76.0 \textsubscript{0.3} & \heatcell{g2}72.8 \textsubscript{0.3} & \heatcell{g2}70.2 \textsubscript{0.5}  & \heatcell{g2}63.2 \textsubscript{0.1}  & \heatcell{g1}61.1 \textsubscript{0.2}  & \heatcell{g2}58.9 \textsubscript{0.6}  & \heatcell{g3}56.5 \textsubscript{0.4}  & \heatcell{g3}54.2 \textsubscript{0.5}
 \\ 
& \textsc{50\%}  & \heatcell{g1}81.2 \textsubscript{0.1} & \heatcell{g1}78.5 \textsubscript{0.2} & \heatcell{g1}75.9 \textsubscript{0.3} & \heatcell{g4}73.1 \textsubscript{0.4} & \heatcell{g4}70.6 \textsubscript{0.5}  & \heatcell{g1}63.1 \textsubscript{0.1}  & \heatcell{g2}61.2 \textsubscript{0.3}  & \heatcell{g2}58.9 \textsubscript{0.6}  & \heatcell{g1}56.2 \textsubscript{0.5}  & \heatcell{g2}54.0 \textsubscript{0.7}
 \\ 
& \textsc{75\%}  & \heatcell{g3}81.5 \textsubscript{0.1} & \heatcell{g4}78.9 \textsubscript{0.3} & \heatcell{g4}76.3 \textsubscript{0.2} & \heatcell{g5}\textbf{73.3} \textsubscript{0.4} & \heatcell{g4}70.6 \textsubscript{0.7} & \heatcell{g3}63.4 \textsubscript{0.1}  & \heatcell{g3}61.4 \textsubscript{0.3}  & \heatcell{g3}59.1 \textsubscript{0.7}  & \heatcell{g4}56.6 \textsubscript{0.5}  & \heatcell{g4}54.4 \textsubscript{0.7}
 \\ 
& \textsc{100\%} & \heatcell{g4}81.7 \textsubscript{0.1} & \heatcell{g5}\textbf{79.0} \textsubscript{0.2} & \heatcell{g5}\textbf{76.4} \textsubscript{0.4} & \heatcell{g5}\textbf{73.3} \textsubscript{0.5} & \heatcell{g5}\textbf{70.8} \textsubscript{0.6}  & \heatcell{g5}\textbf{64.7} \textsubscript{0.0}  & \heatcell{g5}\textbf{62.7} \textsubscript{0.3}  & \heatcell{g5}\textbf{60.3} \textsubscript{0.8}  & \heatcell{g5}\textbf{57.9} \textsubscript{0.3}  & \heatcell{g5}\textbf{55.4} \textsubscript{0.7}
 \\ 

\bottomrule
\end{tabular}%
}
\caption{Results of evaluating the fine-tuned models under various instruction perturbations using the BBH evaluation benchmark. The average exact match (EM) is reported on both the original evaluation instructions (\textsc{0\%}) and the various perturbed evaluation instructions. \textbf{Bold} values denote the best performance across each model.} 
\label{table:Mix_updated_bbh_results}
\end{center}
\end{table*}


\begin{table*}[!t]
\begin{center}
\small
\resizebox{0.83\textwidth}{!}{%
\begin{tabular}{clccccc|ccccc}
\toprule
& & \multicolumn{5}{c}{\textbf{GSM8K (CoT)}} & \multicolumn{5}{c}{\textbf{GSM8K (direct)}} \\

\textbf{} & \textbf{IT} & \textsc{0\%} & \textsc{25\%} & \textsc{50\%} & \textsc{75\%} & \textsc{100\%} & \textsc{0\%} & \textsc{25\%} & \textsc{50\%} & \textsc{75\%} & \textsc{100\%}\\

\midrule

\multirow{6}{*}{\rotatebox{90}{\footnotesize Qwen 7B}}
& \textsc{van.}  & 79.9 \textsubscript{0.2}  & 12.5 \textsubscript{0.3}  & 22.9 \textsubscript{0.6}  & 33.0 \textsubscript{1.4}  & 42.7 \textsubscript{1.2}  & 22.6 \textsubscript{0.0}  & 5.2 \textsubscript{0.3}  & 8.1 \textsubscript{0.6}  & 10.6 \textsubscript{1.0}  & 13.9 \textsubscript{0.6}

 \\ 
 \cmidrule{2-12}
& \textsc{0\%}  & \heatcell{g4}80.6 \textsubscript{0.0} & \heatcell{g3}12.6 \textsubscript{0.5} & \heatcell{g2}24.5 \textsubscript{0.5} & \heatcell{g5}\textbf{34.6} \textsubscript{1.0} & \heatcell{g4}44.6 \textsubscript{1.2}  & \heatcell{g3}25.0 \textsubscript{0.0}  & \heatcell{g3}4.9 \textsubscript{0.3}  & \heatcell{g2}8.3 \textsubscript{0.6}  & \heatcell{g3}11.8 \textsubscript{0.5}  & \heatcell{g3}16.2 \textsubscript{0.9}

 \\ 
& \textsc{25\%}  & \heatcell{g5}\textbf{81.1} \textsubscript{0.0} & \heatcell{g3}12.6 \textsubscript{0.4} & \heatcell{g4}24.7 \textsubscript{0.5} & \heatcell{g4}34.5 \textsubscript{1.2} & \heatcell{g1}44.2 \textsubscript{1.4}  & \heatcell{g2}24.9 \textsubscript{0.0}  & \heatcell{g3}4.9 \textsubscript{0.3}  & \heatcell{g3}8.4 \textsubscript{0.6}  & \heatcell{g3}11.8 \textsubscript{0.4}  & \heatcell{g2}16.1 \textsubscript{1.1}

 \\ 
& \textsc{50\%}  & \heatcell{g4}80.6 \textsubscript{0.0} & \heatcell{g3}12.6 \textsubscript{0.5} & \heatcell{g3}24.6 \textsubscript{0.5} & \heatcell{g3}34.3 \textsubscript{0.7} & \heatcell{g3}44.5 \textsubscript{1.1}  & \heatcell{g4}25.3 \textsubscript{0.0}  & \heatcell{g3}4.9 \textsubscript{0.2}  & \heatcell{g3}8.4 \textsubscript{0.7}  & \heatcell{g2}11.7 \textsubscript{0.4}  & \heatcell{g2}16.1 \textsubscript{0.9}

 \\
& \textsc{75\%}  & \heatcell{g3}80.5 \textsubscript{0.0} & \heatcell{g5}\textbf{12.8} \textsubscript{0.5} & \heatcell{g1}24.4 \textsubscript{0.6} & \heatcell{g2}34.0 \textsubscript{0.9} & \heatcell{g2}44.4 \textsubscript{0.9}  & \heatcell{g2}24.9 \textsubscript{0.0}  & \heatcell{g2}4.8 \textsubscript{0.3}  & \heatcell{g4}8.6 \textsubscript{0.6}  & \heatcell{g5}\textbf{12.1} \textsubscript{0.3}  & \heatcell{g3}16.2 \textsubscript{1.1}
 
 \\ 
& \textsc{100\%} & \heatcell{g2}80.0 \textsubscript{0.0} & \heatcell{g3}12.6 \textsubscript{0.2} & \heatcell{g5}\textbf{24.8} \textsubscript{0.6} & \heatcell{g3}34.3 \textsubscript{1.1} & \heatcell{g5}\textbf{45.1} \textsubscript{0.8}  & \heatcell{g5}\textbf{25.7} \textsubscript{0.0}  & \heatcell{g5}\textbf{5.0} \textsubscript{0.3}  & \heatcell{g5}\textbf{8.7} \textsubscript{0.5}  & \heatcell{g5}\textbf{12.1} \textsubscript{0.3}  & \heatcell{g5}\textbf{16.3} \textsubscript{0.9}

 \\ 

\midrule

\multirow{6}{*}{\rotatebox{90}{\footnotesize Llama 8B}}
& \textsc{van.} & 56.3 \textsubscript{0.3}  & 9.0 \textsubscript{0.7}  & 16.3 \textsubscript{0.6}  & 23.5 \textsubscript{0.6}  & 30.5 \textsubscript{1.2}  & 14.3 \textsubscript{0.1}  & 3.9 \textsubscript{0.3}  & 5.6 \textsubscript{0.7}  & 7.3 \textsubscript{0.6}  & 8.1 \textsubscript{0.5}

 \\ 
 \cmidrule{2-12}
& \textsc{0\%}  & \heatcell{g4}58.4 \textsubscript{0.0} & \heatcell{g3}9.2 \textsubscript{0.1} & \heatcell{g1}16.6 \textsubscript{0.7} & \heatcell{g2}23.8 \textsubscript{1.0} & \heatcell{g5}\textbf{28.1} \textsubscript{1.0}  & \heatcell{g4}14.3 \textsubscript{0.0}  & \heatcell{g3}3.6 \textsubscript{0.2}  & \heatcell{g2}5.0 \textsubscript{0.3}  & \heatcell{g3}6.7 \textsubscript{0.5}  & \heatcell{g3}7.4 \textsubscript{1.0}

 \\ 
& \textsc{25\%} & \heatcell{g5}\textbf{58.5} \textsubscript{0.0} & \heatcell{g5}\textbf{9.4} \textsubscript{0.2} & \heatcell{g2}16.8 \textsubscript{0.8} & \heatcell{g3}23.9 \textsubscript{0.7} & \heatcell{g2}27.7 \textsubscript{0.9} & \heatcell{g5}\textbf{14.6} \textsubscript{0.0} & \heatcell{g5}\textbf{3.9} \textsubscript{0.1}  & \heatcell{g2}5.0 \textsubscript{0.4}  & \heatcell{g2}6.6 \textsubscript{0.4}  & \heatcell{g2}7.3 \textsubscript{1.0}

 \\ 
& \textsc{50\%}  & \heatcell{g3}58.2 \textsubscript{0.0} & \heatcell{g3}9.2 \textsubscript{0.2} & \heatcell{g3}16.9 \textsubscript{0.6} & \heatcell{g5}\textbf{24.0} \textsubscript{0.9} & \heatcell{g3}27.8 \textsubscript{1.0} & \heatcell{g2}14.1 \textsubscript{0.0}  & \heatcell{g2}3.5 \textsubscript{0.1}  & \heatcell{g2}5.0 \textsubscript{0.4}  & \heatcell{g2}6.6 \textsubscript{0.4}  & \heatcell{g4}7.5 \textsubscript{1.1}

 \\
& \textsc{75\%} & \heatcell{g1}57.4 \textsubscript{0.0} & \heatcell{g1}9.1 \textsubscript{0.2} & \heatcell{g3}16.9 \textsubscript{0.7} & \heatcell{g2}23.8 \textsubscript{0.8} & \heatcell{g1}27.6 \textsubscript{1.2} & \heatcell{g2}14.1 \textsubscript{0.0}  & \heatcell{g4}3.7 \textsubscript{0.1}  & \heatcell{g3}5.2 \textsubscript{0.4}  & \heatcell{g2}6.6 \textsubscript{0.4}  & \heatcell{g4}7.5 \textsubscript{1.1}

 \\ 
& \textsc{100\%} & \heatcell{g4}58.4 \textsubscript{0.0} & \heatcell{g3}9.2 \textsubscript{0.2} & \heatcell{g5}\textbf{17.1} \textsubscript{0.6} & \heatcell{g1}23.7 \textsubscript{1.2} & \heatcell{g3}27.8 \textsubscript{1.5} & \heatcell{g3}14.2 \textsubscript{0.0}  & \heatcell{g3}3.6 \textsubscript{0.0}  & \heatcell{g5}\textbf{5.3} \textsubscript{0.3}  & \heatcell{g5}\textbf{6.9} \textsubscript{0.3}  & \heatcell{g5}\textbf{7.8} \textsubscript{1.0}

 \\ 

\midrule

\multirow{6}{*}{\rotatebox{90}{\footnotesize Qwen 72B}}
& \textsc{van.}  & 88.8 \textsubscript{0.2}  & 14.9 \textsubscript{0.5}  & 28.1 \textsubscript{0.7}  & 40.8 \textsubscript{1.0}  & 53.0 \textsubscript{1.5}  & 43.2 \textsubscript{0.0}  & 7.9 \textsubscript{0.3}  & 14.0 \textsubscript{0.7}  & 19.7 \textsubscript{1.0}  & 25.5 \textsubscript{1.0}

 \\ 
 \cmidrule{2-12}
& \textsc{0\%}  & \heatcell{g3}90.0 \textsubscript{0.2} & \heatcell{g2}15.3 \textsubscript{0.4} & \heatcell{g1}29.0 \textsubscript{0.4} & \heatcell{g1}42.7 \textsubscript{1.1} & \heatcell{g1}55.0 \textsubscript{1.7} & \heatcell{g5}\textbf{44.8} \textsubscript{0.2}  & \heatcell{g2}8.3 \textsubscript{0.7}  & \heatcell{g3}15.3 \textsubscript{0.8}  & \heatcell{g3}21.3 \textsubscript{0.9}  & \heatcell{g5}\textbf{27.6} \textsubscript{1.0}
 \\ 
& \textsc{25\%}  & \heatcell{g1}89.9 \textsubscript{0.2} & \heatcell{g2}15.3 \textsubscript{0.5} & \heatcell{g4}29.6 \textsubscript{0.4} & \heatcell{g4}43.0 \textsubscript{1.1} & \heatcell{g3}55.5 \textsubscript{1.6} & \heatcell{g4}44.7 \textsubscript{0.1}  & \heatcell{g3}8.4 \textsubscript{0.6}  & \heatcell{g2}15.1 \textsubscript{0.5}  & \heatcell{g2}21.1 \textsubscript{0.7}  & \heatcell{g3}27.5 \textsubscript{1.0}
 \\ 
& \textsc{50\%} & \heatcell{g1}89.9 \textsubscript{0.2} & \heatcell{g3}15.4 \textsubscript{0.4} & \heatcell{g2}29.3 \textsubscript{0.5} & \heatcell{g2}42.8 \textsubscript{1.4} & \heatcell{g3}55.5 \textsubscript{1.9} & \heatcell{g3}44.2 \textsubscript{0.1}  & \heatcell{g2}8.3 \textsubscript{0.5}  & \heatcell{g2}15.1 \textsubscript{0.4}  & \heatcell{g5}\textbf{21.4} \textsubscript{0.6}  & \heatcell{g3}27.5 \textsubscript{0.9}
 \\ 
& \textsc{75\%}  & \heatcell{g5}\textbf{90.3} \textsubscript{0.3} & \heatcell{g5}\textbf{15.6} \textsubscript{0.3} & \heatcell{g4}29.6 \textsubscript{0.7} & \heatcell{g3}42.9 \textsubscript{1.7} & \heatcell{g3}55.5 \textsubscript{2.3} & 4\heatcell{g1}3.7 \textsubscript{0.1}  & \heatcell{g5}\textbf{8.5} \textsubscript{0.6}  & \heatcell{g5}\textbf{15.4} \textsubscript{0.5}  & \heatcell{g5}\textbf{21.4} \textsubscript{0.6}  & \heatcell{g5}\textbf{27.6} \textsubscript{0.8}
 \\ 
& \textsc{100\%} & \heatcell{g4}90.2 \textsubscript{0.1} & \heatcell{g5}\textbf{15.6} \textsubscript{0.5} & \heatcell{g5}\textbf{29.7} \textsubscript{0.7} & \heatcell{g5}\textbf{43.4} \textsubscript{1.8} & \heatcell{g5}\textbf{55.9} \textsubscript{2.0} & \heatcell{g2}44.0 \textsubscript{0.2}  & \heatcell{g3}8.4 \textsubscript{0.5}  & \heatcell{g3}15.3 \textsubscript{0.5}  & \heatcell{g5}\textbf{21.4} \textsubscript{0.7}  & \heatcell{g3}27.5 \textsubscript{1.0}
 \\ 

\midrule

\multirow{6}{*}{\rotatebox{90}{\footnotesize Llama 70B}}
& \textsc{van.}  & 80.2 \textsubscript{0.1}  & 13.0 \textsubscript{0.3}  & 24.1 \textsubscript{0.5}  & 34.7 \textsubscript{1.5}  & 43.9 \textsubscript{0.9}  & 34.4 \textsubscript{0.2}  & 6.6 \textsubscript{0.6}  & 10.9 \textsubscript{0.6}  & 14.7 \textsubscript{1.0}  & 19.1 \textsubscript{1.2}
 \\ 
 \cmidrule{2-12}
& \textsc{0\%}   & \heatcell{g5}\textbf{82.3} \textsubscript{0.2} & \heatcell{g3}13.7 \textsubscript{0.7} & \heatcell{g2}25.4 \textsubscript{0.4} & \heatcell{g2}37.0 \textsubscript{1.0} & \heatcell{g1}47.5 \textsubscript{1.0} & \heatcell{g5}\textbf{35.3} \textsubscript{0.1}  & \heatcell{g2}6.8 \textsubscript{0.5}  & \heatcell{g3}11.8 \textsubscript{0.4}  & \heatcell{g2}16.2 \textsubscript{0.6}  & \heatcell{g2}20.7 \textsubscript{0.8}
 \\ 
& \textsc{25\%}  & \heatcell{g4}82.1 \textsubscript{0.2} & \heatcell{g5}\textbf{14.0} \textsubscript{0.6} & \heatcell{g3}25.5 \textsubscript{0.8} & \heatcell{g2}37.0 \textsubscript{1.3} & \heatcell{g3}47.7 \textsubscript{0.5} & \heatcell{g4}35.0 \textsubscript{0.1}  & \heatcell{g3}7.0 \textsubscript{0.5}  & \heatcell{g2}11.7 \textsubscript{0.7}  & \heatcell{g3}16.3 \textsubscript{0.7}  & \heatcell{g3}20.8 \textsubscript{0.8}
 \\ 
& \textsc{50\%} & \heatcell{g1}80.2 \textsubscript{0.3} & \heatcell{g2}13.5 \textsubscript{0.6} & \heatcell{g4}25.6 \textsubscript{0.7} & \heatcell{g2}37.0 \textsubscript{1.4} & \heatcell{g2}47.6 \textsubscript{1.0} & \heatcell{g1}34.3 \textsubscript{0.2}  & \heatcell{g3}7.0 \textsubscript{0.5}  & \heatcell{g3}11.8 \textsubscript{0.8}  & \heatcell{g4}16.4 \textsubscript{0.5}  & \heatcell{g3}20.8 \textsubscript{0.8}
 \\ 
& \textsc{75\%}  & \heatcell{g2}81.6 \textsubscript{0.2} & \heatcell{g4}13.8 \textsubscript{0.4} & \heatcell{g4}25.6 \textsubscript{0.7} & \heatcell{g3}37.3 \textsubscript{1.1} & \heatcell{g4}48.2 \textsubscript{0.7} & \heatcell{g2}34.4 \textsubscript{0.1}  & \heatcell{g5}\textbf{7.1} \textsubscript{0.6}  & \heatcell{g4}11.9 \textsubscript{0.5}  & \heatcell{g5}\textbf{16.6} \textsubscript{0.9}  & \heatcell{g3}20.8 \textsubscript{0.8}
 \\ 
& \textsc{100\%}  & \heatcell{g3}82.0 \textsubscript{0.2} & \heatcell{g3}13.7 \textsubscript{0.4} & \heatcell{g5}\textbf{26.1} \textsubscript{0.6} & \heatcell{g5}\textbf{38.1} \textsubscript{1.8} & \heatcell{g5}\textbf{48.8} \textsubscript{1.2} & \heatcell{g3}34.5 \textsubscript{0.2}  & \heatcell{g3}7.0 \textsubscript{0.5}  & \heatcell{g5}\textbf{12.1} \textsubscript{0.9}  & \heatcell{g4}16.4 \textsubscript{0.6}  & \heatcell{g5}\textbf{21.4} \textsubscript{0.7}
 \\ 

\bottomrule
\end{tabular}%
}
\caption{Results of evaluating the fine-tuned models under various instruction perturbations using the GSM8K evaluation benchmark. The exact match (EM) is reported on both the original evaluation instructions (\textsc{0\%}) and the various perturbed evaluation instructions. \textbf{Bold} values denote the best performance across each model.} 
\label{table:Mix_updated_GSM_results}
\end{center}
\end{table*}

\end{document}